\documentclass[nopreprintline,12pt,authoryear]{elsarticle}

\usepackage{amssymb}
\usepackage[utf8]{inputenc} 
\usepackage[T1]{fontenc}    
\usepackage{url}            
\usepackage{booktabs}       
\usepackage{amsfonts}       
\usepackage{nicefrac}       
\usepackage{microtype}      
\usepackage{xcolor}         
\usepackage[hidelinks]{hyperref}
\usepackage[cmex10]{amsmath}
\usepackage{amsthm}
\usepackage{algorithm,tabularx} 
\usepackage{algpseudocode}
\usepackage{xspace}

\usepackage{caption}
\usepackage{subcaption}
\usepackage{varwidth}
\usepackage{graphicx}
\usepackage{multirow}

\newcommand{\CDKT}{\ensuremath{\textsf{CDKT-FL}}\xspace}
\journal{Nuclear Physics B}

\begin{document}

\begin{frontmatter}


\fntext[fn1]{Equal contribution.}

\cortext[cor1]{Corresponding author.}


\title{CDKT-FL: Cross-Device Knowledge Transfer using Proxy Dataset in Federated Learning}


\author[label1]{Huy Q. Le \fnref{fn1}}
\ead{quanghuy69@khu.ac.kr}
\author[label3]{ Minh N. H. Nguyen \fnref{fn1}}
\ead{nhnminh@vku.udn.vn}
\author[label4]{Shashi Raj Pandey}
\ead{srp@es.aau.dk}
\author[label2]{Chaoning Zhang}
\ead{chaoningzhang1990@gmail.com}
\author[label1]{Choong Seon Hong \corref{cor1}}
\ead{cshong@khu.ac.kr}

\affiliation[label1]{organization={Department of Computer Science and Engineering, Kyung Hee University},
            city={Yongin-si},
            postcode={17104}, 
            country={Republic of Korea}}
\affiliation[label2]{organization={Department of Artificial Intelligence, Kyung Hee University},
            city={Yongin-si},
            postcode={17104}, 
            country={Republic of Korea}}
\affiliation[label3]{organization={Digital Science and Technology Institute, The University of Danang—Vietnam-Korea University of Information and Communication Technology},
            city={Da Nang},
            postcode={550000}, 
            country={Vietnam}}
\affiliation[label4]{organization={Department of Electronic Systems, Aalborg University},
            city={Aalborg},
            postcode={9220}, 
            country={Denmark}}

\begin{abstract}
In a practical setting, how to enable robust Federated Learning (FL) systems, both in terms of generalization and personalization abilities, is one important research question. It is a challenging issue due to the consequences of non-i.i.d. properties of client's data, often referred to as statistical heterogeneity, and small local data samples from the various data distributions. Therefore, to develop robust generalized global and personalized models, conventional FL methods need to redesign the knowledge aggregation from biased local models while considering huge divergence of learning parameters due to skewed client data. In this work, we demonstrate that the knowledge transfer mechanism achieves these objectives and develop a novel knowledge distillation-based approach to study the extent of knowledge transfer between the global model and local models. Henceforth, our method considers the suitability of transferring the outcome distribution and (or) the embedding vector of representation from trained models during cross-device knowledge transfer using a small proxy dataset in heterogeneous FL. In doing so, we alternatively perform cross-device knowledge transfer following general formulations as 1) global knowledge transfer and 2) on-device knowledge transfer. \textcolor{black}{Through simulations on three federated datasets, we show the proposed method achieves significant speedups and high personalized performance of local models. Furthermore, the proposed approach offers a more stable algorithm than other baselines during the training, with minimal communication data load when exchanging the trained model's outcomes and representation.}
\end{abstract}




\end{frontmatter}


\section{Introduction}
Nowadays, myriad mobile applications incorporate on-device AI modules to utilize the user's data and facilitate new experiences on their smartphones, such as AI-powered cameras, Extended Reality (XR) applications, and intelligent assistants \citep{onDevSamsung}. However, data privacy concerns have been an \textit{Achille's heels} in the centralized machine learning (ML) paradigm. Thus, in recent years, there has been a sudden surge in developing federated learning (FL) frameworks that are able to limit data privacy leakage~\citep{FL_advances,mcmahan2016communication,federatedSurvey2020,li2021model,nguyen2021distributed,nguyen2022self,thwal2024ondev}. In FL, a set of devices are connected with a central node (edge/cloud server) to collaboratively train a learning model without sharing their local data \citep{FL_advances}. In doing so, the FL schemes require sharing only the learning model parameters with the server; hence, the user data privacy can be kept intact. However, such a federated training scheme raises concerns about the heterogeneity of computing devices and distributed data samples' properties, leading to poor learning performance and fluctuation of the trained models across devices. To fill this research gap, recent FL algorithm designs are departing from heterogeneous FL settings towards personalized FL, which considers the performance of the global model and the personalized client models. In \textit{heterogeneous} FL, the underlying data distribution is not identical across devices, i.e., non-i.i.d. Hence, the solution to the global empirical risk minimization problem, i.e., the trained global model, will not \textit{personalize} well for each client. 
Meanwhile, the local training process in FL algorithms with their small private datasets often results in biased client models that produce high accuracy on the local data but with low generalization capabilities. Also, the client model parameters and data features are diverse across devices. Such as in FedAvg~\citep{mcmahan2016communication}, the aggregation of model parameters based on the conventional coordinates-based averaging scheme from highly skewed data could slow down the learning process due to extensive divergence of model parameters to be averaged \citep{yu2020heterogeneous}.

In this paper, we redesign knowledge transfer mechanisms in FL and examine the underlying fundamental correlation between the learning performances of FL algorithms. \textcolor{black}{Unlike previous FL methods~\citep{mcmahan2016communication, FedProx2020, acar2021federated, guzzo2023data} that adopt the parameter aggregation scheme, our framework introduces the knowledge transfer mechanism on both sides to update the models without exchanging weight parameters of learning models between server and local devices.
Instead of sending parameters, the models only need to exchange knowledge, such as the outcomes or representations. Specifically, we propose \CDKT, a novel cross-device knowledge transfer mechanism that leverages the model outcomes and (or) representation from clients using small proxy data to achieve robust generalized global and personalized models.} To that end, we show that the proposed method achieves significant speedups and obtains better personalized performance of local models in the fixed users scenario and the subset of users selection scenario. \textcolor{black}{Furthermore, we show that our approach offers better stability than other baselines and tackles privacy leakage issues with minimal communication data load when exchanging the trained model's outcomes and (or) representation solely. Our main contributions are:
\begin{itemize}
    \item In the typical FL algorithms such as FedAvg \citep{mcmahan2016communication}, FedProx \citep{FedProx2020}, the global model is updated via model aggregation (i.e., parameters averaging), which is then broadcast back to clients and used as the initial model for local learning. However, such approaches ignore the fundamentals of interleaved properties of data across clients, thus resulting in poor generalization. To address this research gap, in this work, we propose cross-device knowledge transfer (CDKT) with two mechanisms: 1) Generalized Model Construction for transferring knowledge from client models to the global model, and 2) On-Device Learning for transferring the generalized knowledge of global model to client models. 
    \item In doing so, we study the knowledge transfer mechanism using generic distance regularizers to align the knowledge between the models. Unlike knowledge distillation approaches, the transferable knowledge is the outcomes (i.e., activation results $z$ of the final layer) and (or) embedding representation features (i.e., activation results $e$ of the intermediate layer) given the proxy samples. We assume this proxy data is of small size and available by clients to follow a practical scenario where the system can pre-collect a small amount of labeled data at the beginning.
    \item Our experiments reveal that (i) the outcome distribution and embedding features' representation of the proxy data shared by all devices and servers are transferable and (ii) quickly improves the global model performance and personalized performance of the client models. Unlike many existing works, we evaluate the personalized performance of local models with specialization (i.e., performance on local testing data), and generalization capabilities (i.e., cross-device testing performance). These evaluation metrics provide more insights into the behaviors of the heterogeneous FL systems with different transferring methods and learning settings.
    \item  The proposed \CDKT algorithm enables the flexibility of heterogeneous learning model design and knowledge transfer mechanism for future personalized FL applications. CDKT also improves stability and faster convergence while requiring fewer communication data with the outcome and representation knowledge than full complex model parameters compared to the FedAvg algorithm. These observations provide solid motivation to study knowledge transfer approaches and related metrics in FL. To the best of our knowledge, even though there are existing KD-based algorithms, our general proposed knowledge transfer approach and metrics advance the applicability of KD-based algorithms. Further, different from existing works, we leverage the true labels that can fix the wrong prediction from weak models in the early stage of the training process and improve the generalization of models.
\end{itemize}
}
\textcolor{black}{To that end, we discuss some related works in Section \ref{s:relatedwork}. We present our proposed method and formulate generalized model construction, on-device learning and propose a \CDKT algorithm in Section \ref{s:method}. We validate the effectiveness of our proposed algorithm for both specialization and generalization performance of the client and global models compared to several state-of-the-art FL algorithms in Section \ref{s:experiment}. Finally, Section \ref{s:conclusion} concludes the paper.}

\section{Related Works}\label{s:relatedwork}
\textbf{Structural alignment in heterogeneous FL:} One of the promising research directions to resolve unmatched parameters of learning models in FL is using the structural alignment methods \citep{mostafa2019robust,wang2020federated,yu2020heterogeneous,zhang2021federated,sun2023feature,Xiong_2023_CVPR}. The representation matching in \citep{mostafa2019robust} and ``permutation invariance" in FedMA \citep{wang2020federated} are proposed to evaluate parameter similarity across local models and reorganize the permutation accordingly via matching schemes; thus, obtaining parameters that could be averaged together in the model aggregation. In \citep{yu2020heterogeneous}, the authors proposed structure-information alignment based on the feature-oriented interpretation and regulation method to ensure explicit feature information allocation in different neural network structures. Based on the feature interpretation, the proposed regulation method adjusts the local model architecture according to their data and task distribution at the very early training stage and continuously maintains structure and information alignment. To address the domain generalization problem in FL, the authors in~\citep{zhang2021federated} design the federated adversarial learning approach to measure and align the distributions among different source domains by matching each distribution to a reference distribution in a distributed manner. Similarly,~\citet{sun2023feature} tackle the domain generalization problem using feature distribution matching, which employs a federated voting mechanism to generate pseudo-labels for the target domain by aggregating the consensus of clients to fine-tune a global model. Recently,~\citet{Xiong_2023_CVPR} proposed the FedDM method, which utilizes iterative distribution matching to learn a surrogate function. Instead of sending local model updates, FedDM sends synthesized data to the server, resulting in improved communication efficiency and effectiveness. In this manner, structural alignment methods showed their efficiency in improving the FL performance and applicability in various heterogeneous settings.

\textbf{Common Representation Learning:}
This is a popular method to handle the uneven distributions and representations of different modalities or tasks \citep{crosstask2013,huang2017cross,peng2019cm,xu2020learning}. In this approach, the training of shared features across tasks/modalities constructs a common representation; thereby, cross-modal common representational learning allows knowledge transfer across different modalities \citep{crosstask2013}. In doing so, the heterogeneous data from different modalities resulting in diverse representations are projected into a common space. Then, the cross-modal knowledge transfer can be performed by reducing the discrepancy between representations of pairwise cross-modal data from high-level layers \citep{huang2017cross}. Furthermore, it can preserve the relevant semantics and allow heterogeneous data from different modalities to be correlated more easily \citep{peng2019cm,xu2020learning}. Recently, a similar design, the FedRep algorithm \citep{collins2021exploiting}, was applied in FL that alternatively trains the shared low-dimensional representation, which is aggregated across clients by averaging and private local heads to improve the learning performance of client models. In contrast to FedRep, the approach presented in~\citep{oh2022fedbabu} introduces FedBABU, an algorithm that focuses on learning and aggregating the body of the model while keeping the head randomly initialized during local updates.

\textbf{The convergence of the alignment methods and knowledge distillation:}
Even though the alignment approaches are widely studied in the representation learning literature, the knowledge distillation (KD) approach has emerged as one of the promising alignment approaches. KD approach demonstrated superior performance via knowledge transfer using outcome distribution (i.e., soft targets) from the teacher to student models~\citep{hinton2015distilling}. The alignment of outcome distribution using Kullback–Leibler (KL) divergence helps the student model imitate and learn from a teacher model or ensemble of models. KD was first introduced in~\citep{bucilua2006model,ba2013deep} and then developed by \citep{hinton2015distilling} to compress the smaller learning model from a large model. In KD, the soft targets $z_t$ and $z_s$ are the softmax outputs of the student and teacher model, respectively. The loss function of the student model between the prediction $z_s$ and ground-truth label $y_s$ is the cross-entropy loss denoted as $\ell_{CE}$. In addition, the distillation loss using the KL divergence is defined as

\begin{equation}
    \ell_{KD}=\tau^2 KL(z_t, z_s),
\end{equation}
where $\tau$ is the temperature parameter that softens the output values. The higher the value of the temperature, the softer the outputs of learning models. Therefore, the loss function is used to train the student model is defined as 

\begin{equation}
    \ell_{\textrm{student}} = \ell_{CE} + \alpha\ell_{KD},
\end{equation}
where $\alpha$ is the hyper-parameter that controls the penalty KD loss. \\

In \citep{zhang2018deep}, the authors introduced a deep mutual learning strategy with an ensemble of student models to collaboratively learn and teach each other throughout the training process. Furthermore, KD was recently adopted to transfer the knowledge between the global and local models and exhibited promising results in FL. The alignment of outcome distribution with KL divergence in KD helps the global model and local models to transfer their learning knowledge and further improve the performance of the learning models. To overcome communication cost and heterogeneity issues, the authors in \citep{jeong2018communication} applied KD with the average logits from clients using their private datasets. Balancing the private datasets with data augmentation can further improve the learning performance. The works \citep{li2019fedmd,bistritz2020distributed,9562670} leveraged a large public dataset in the local knowledge distillation using the aggregation/consensus of soft targets from clients as the teacher knowledge. In \citep{lin2020ensemble}, the authors proposed ensemble distillation for global model fusion, enabling different sizes, numerical precision, or structure of learning models. It is possible to apply KD in the global and local model updates, as shown in \citep{he2020group}, but the server requires sequentially distilling the knowledge from each client and vice versa. However, the authors do not consider the proxy dataset as well as the knowledge aggregation. Recently, the authors in~\citep{zhang2021parameterized} introduced the personalized knowledge transfer mechanism for personalized FL utilizing proxy data to update the local soft predictions for each local client. In addition to utilizing proxy data for knowledge aggregation from local models,~\citep{zhu2021data,Zhang_2022_CVPR} employs a lightweight generator to combine the knowledge of client models in a data-free manner. \textcolor{black}{Moreover, another innovative approach known as split learning has garnered attention in federated learning research~\citep{vepakomma2018split,singh2019detailed,thapa2022splitfed}. Unlike traditional FL methods, split learning employs a strategy where knowledge is shared using activations and gradients, ensuring privacy preservation in collaborative learning. Split learning primarily involves the division of the neural network model into two segments: device models for each device and a central server model that receives activations from the devices as input. The process includes transmitting activations, referred to as smashed data, from the split layer, known as the cut layer, of the client-side network to the server and receiving the gradients of the smashed data resulting from server-side operations. Unlike split learning, CDKT has the global learning model having generalized capabilities and playing a role in knowledge aggregation from client models while preserving the good personalized performance of local models leveraging proxy data.}

Inspired by the discussion on alignment methods and knowledge distillation approaches, it is fundamental to find the answer to the following research question: 
``\textit{what transferable knowledge and regularizers should be used in knowledge transfer between the global model and local models in FL?}" Answering this question opens opportunities to provide more efficient knowledge transfer mechanisms in FL. 
\begin{figure*}[t]
	\centering
	\includegraphics[width=0.9\linewidth]{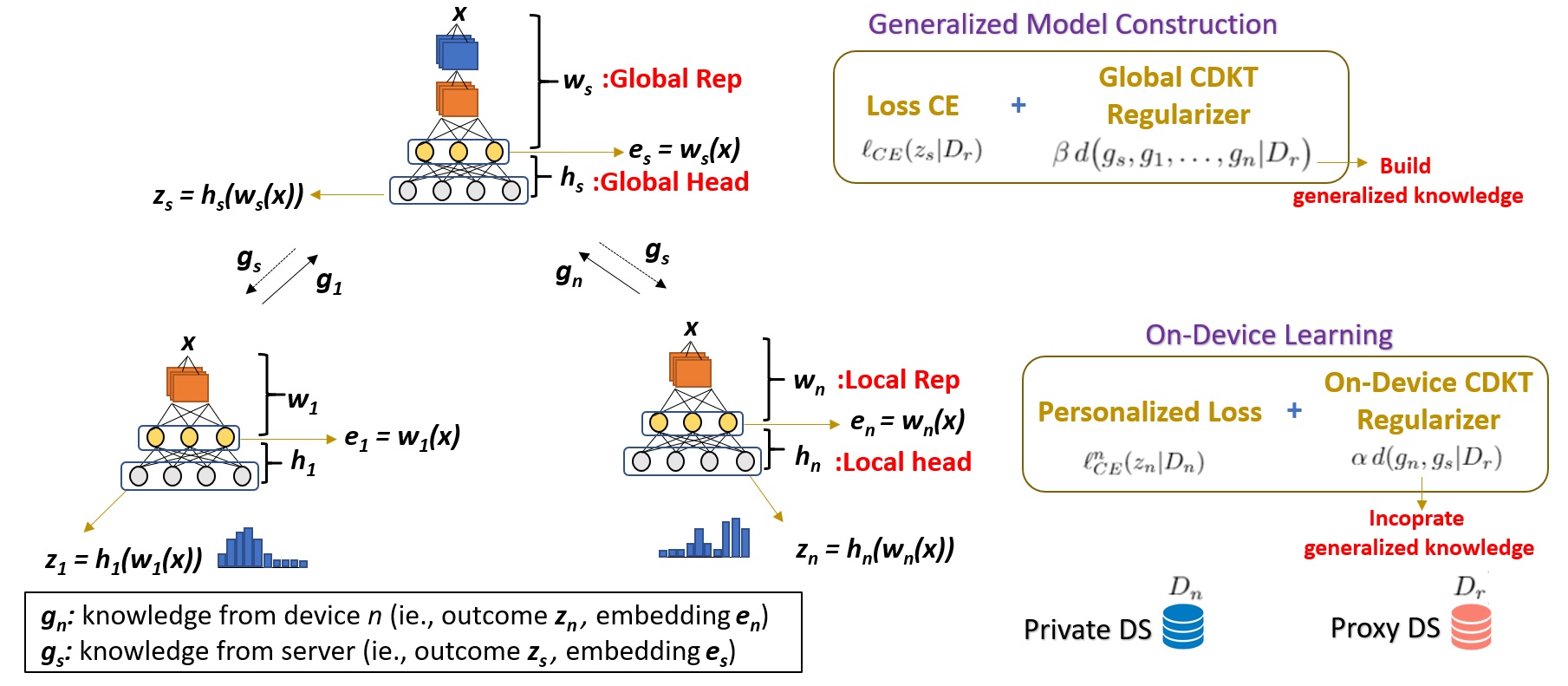}
	\caption{\textcolor{black}{An overview of Cross-Device Knowledge Transfer (CDKT) scheme in FL, integrating two mechanisms: 1) Generalized Model Construction for transferring knowledge from client models to the global model, and 2) On-Device Learning for transferring the generalized knowledge of global model to client models.}}
	\label{F:Scheme}
\end{figure*}
\section{Cross-Device Knowledge Transfer in Federated Learning}\label{s:method}

In a general scheme of FL algorithms \citep{reddi2020adaptive,li2021ditto}, the global model is updated via a gradient-based optimizer or model parameter aggregation in the server, while the local/personalized model in user device $n$ is updated using their private dataset $D_n$. We first define the neural network with the prediction outcomes (i.e., $z=h\circ w(x)$), where $h$ is the projection head, which is a small neural network (i.e., the last fully connected layers; a small classifier). We denote the embedding features of the representation sub-net parameters $w$ for the image $x$ as $e=w(x)$. Hence, the learning model parameters $w, h$ can be trained and used to extract the embedding features (i.e., $e_n$) and (or) classes' prediction distribution (i.e., $z_n$) of input images at device $n$. 
In this work, instead of using model parameters from clients in every round to aggregate into the global model, cross-device knowledge transfer is being used to transfer the collective knowledge from all clients to the server and construct a generalized model. We illustrate the overview of our proposed CDKT-FL scheme in Fig.~\ref{F:Scheme}. Note that the global model can be updated using an additional regularizer to align its knowledge with the collective knowledge of clients. In doing so, it is necessary to have a fixed small proxy data $D_r$ such that all models can access and provide their learning knowledge as the outcomes (i.e., $z$) and (or) embedding features (i.e., $e$). In addition, the knowledge transfer also helps the global model transfer its generalization capability to all client models. This way, firstly, we can reduce communication data load and also further enhance users' data privacy, which is often compromised during full parameters exchange against reverse engineering tricks. Secondly, since knowledge transfer is sufficient to transfer learning capabilities, the learning model architectures can differ between the server model (with more layers) and client models (with fewer layers). 
\begin{algorithm}[t]
	\caption{\CDKT Algorithm} 
	\begin{algorithmic}[1]
    \State \textbf{Input:} \textcolor{black}{Number of global rounds $T$, number of local epochs $K$ and number of global epochs $R$ for global model update}
    \For{ $t=0,\dots,T-1$}
    \State \textbf{On-Device Learning:} Each device $n$ receives the generalized knowledge $\boldsymbol{g}^{}_{s}$ from the server and updates local model with $D_n$, $D_r$;
    \For { $k=0,\dots,K-1$}
        \State \begin{varwidth}[t]{0.85\linewidth}We loop for each batch of private and proxy data (i.e., $S_n$ and $S_r$): \end{varwidth}
         \begin{align}
            {w}^{t}_{n} = &{w}^{t}_{n}-\eta\nabla \ell_c^n(S_n,S_r);
        \end{align}
    \EndFor
    \State \begin{varwidth}[t]{1.\linewidth} Device $n$ extracts and sends the local knowledge $\boldsymbol{g}_n$ given the proxy dataset to the server;
    \end{varwidth}
    \State \textbf{Generalized Model Construction:} Server updates the global model with the  received local knowledge from the selected devices:
     

    \For { $r=0,\dots,R-1$}
        \State We loop through batches of proxy data (i.e., $S_r$):
      \begin{align}
              {w}^{t}_{g}= &{w}^{t}_{g}-\gamma\nabla\ell_s({S}_{r})
     \end{align}
     \EndFor
    \EndFor
    \end{algorithmic}
    \label{alg}
\end{algorithm}	

The proposed \CDKT algorithm (Alg. \ref{alg}) enables the knowledge transfer between clients and global models in the FL setting. At each global round $t$, the server broadcasts the knowledge from the updated global model to all clients. Then, each client performs $K$ local epochs with its private and proxy dataset based on the gradient of the loss function in the \textit{on-device learning} problem (\ref{eq:on-dev-CDKT}), with the local learning rate $\eta$. Note that we use mini-batch of private data in the cross-entropy loss, whereas proxy data is in the on-device CDKT regularizer. Then, the clients send their latest knowledge (i.e., $z_n$ and (or) $e_n$) to the server. At the global level, the server builds the generalized model by solving the global learning problem (\ref{eq:global-CDKT}). For this, we use a gradient-based update for $R$ global epochs with the global learning rate $\gamma$. In particular, to perform \CDKT, we design the knowledge transfer in FL with two general problems: \textit{generalized model construction} and \textit{on-device learning} in the following subsections.


\subsection{Generalized Model Construction}
In the generalized model construction problem, the main target is utilizing the proxy data (i.e., $D_r$) and collective knowledge from all devices to build the global model. If the transferable knowledge is only the outcome of client models, then the problem is in the form of knowledge distillation (KD) \citep{hinton2015distilling}. 
\textcolor{black}{Similar to the design of \citep{li2019fedmd}, we consider the averaged results of the outcomes as the collective knowledge that results in the equivalent performance of the learning problem.} This approach, however, requires less computation than the sum of individual KL loss between the student model (global model) and each teacher model (client model) in ensemble KD. In the following, we review the KD method and discuss the proposed cross-device knowledge transfer method.

\textbf{ \textit{KD Method:}}
\begin{equation}
    {\ell}^{}_{s}(D_r)= {\ell}_{CE}(z_{s}|D_r)+\beta \, {\ell}_{KD}\bigg(z_s,\sum_{n\in{N}}{\frac{1}{N}}z_{n}|D_r\bigg),
\end{equation}
where $\beta$ is the parameter to control the trade-off between proxy data loss and the KD regularizer; $z$ are the soft targets of the learning models.

\textbf{\textit{Cross-device knowledge transfer Method (CDKT):}}
We formulate the generalized model construction problem with the global CDKT regularizer using a generic function $d$. Thereby, the proposed formulation measures the discrepancy between the knowledge of $n$ client models ($g_{1},\dots,g_n$) and the knowledge of server model $g_s$ as
\begin{equation}
    {\ell}^{}_{s}(D_r)= {\ell}_{CE}(z_{s}|D_r)+\beta \, d\big(g_s,g_{1},\dots,g_n |D_r\big).
\end{equation}
The collective knowledge from clients are constructed as the averaged results of the outcomes $z$ and (or) embedding features $e$. To improve the limitation in transferred knowledge from clients, we adjust the collective outcomes $z$ of clients by averaging it with the true labels $y_r$ when using their final outcomes.
\begin{align}
    {\ell}^{}_{s}(D_r)&= {\ell}_{CE}(z_{s}|D_r)+\beta \, d\bigg(e_s,\sum_{n\in{N}}{\frac{1}{N}}e_{n} |D_r\bigg) \notag \\
    &+ \beta \, d\bigg(z_s,\lambda y_r+(1-\lambda)\sum_{n\in{N}}{\frac{1}{N}}z_{n} |D_r\bigg),
    \label{eq:global-CDKT}
\end{align}
where $\lambda\in[0,1]$, and the parameter $\beta$ allows the server to control the trade-off between proxy data loss and the global CDKT regularizer. \textcolor{black}{Here, we note that various distance functions $d$ can be used for knowledge transfer, such as Norm, KL divergence, and Jensen-Shannon (JS) divergence \citep{kullback1997information}}. A generic distance function $d$ in CDKT regularizer can be adopted with the three following distance metrics.

\textbf{Norm2 distance:} $$d_{norm}(g_n,g_s|D_r) = \sum_{x \in D_r} \|g_n(x)-g_s(x)\|_2,$$
where $g_n(x)$ is the prediction outcome (or embedding features) of client $n$ and $g_s(x)$ is the knowledge from the global model given data sample $x$.

\textbf{Kullback–Leibler (KL) divergence:} $$d_{KL}(g_n,g_s|D_r) =   \sum_{x \in D_r}\sum_{c}P(g_{s,c}|x) \frac{P(g_{s,c}|x)}{P(g_{n,c}|x)},$$
where $P(g_{n,c}|x)$ is the prediction outcome (or embedding features) of client $n$ and $P(g_{s,c}|x)$ is the knowledge from the global model of class $c$ given data sample $x$.

\textbf{Jashon-Shannon (JS) divergence:}
$$d_{JS}(g_n,g_s|D_r) =  \frac{1}{2}d_{KL}(g_n, g_m|D_r) + \frac{1}{2}d_{KL}(g_s, g_m|D_r) $$
where $g_m=\frac{1}{2}(g_n+g_s)$.

\textcolor{black}{In this work, we explored a range of distance functions as regularizers to optimize the performance of our collaborative learning framework. First, we aim to identify the most effective regularizers for achieving stable performance in global and local contexts. By experimenting with different regularizers, we aimed to fine-tune the model's behavior to suit the specific demands of each scenario, ensuring optimal performance on a global scale while also accommodating the unique characteristics of local clients. Second, it is essential to note that our collaborative learning framework is flexible and adaptable. Note that the knowledge transfer that is reflected in the regularizer of the global and local learning objectives plays a vital role in the CKDT algorithm to exchange the learning capabilities between models.}
\subsection{On-device Learning}

Using a similar design, the on-device learning problem helps the client models to improve their generalization capabilities by imitating the generalized knowledge from the global model. In this approach, the client $n$ utilizes the private dataset $D_n$ and also the generalized knowledge  obtained from the global model using the proxy dataset $D_r$. In particular, we formulate the on-device loss function for each device $n$ as follows:
\begin{align}
    {\ell}^{n}_{c} (D_n, D_r)
    &= {\ell}^{n}_{CE}(z_n|D_n) + \alpha \, d(g_n,g_s|D_r) \nonumber \\
    &= {\ell}^{n}_{CE}(z_n|D_n) + \alpha \, d(e_n,e_s|D_r) 
    +\alpha \,d(z_n,\lambda y_s + (1-\lambda) z_s|D_r)
    \label{eq:on-dev-CDKT}
\end{align}
where $\alpha$ is the parameter to control the trade-off between the local learning objective's biased knowledge and generalization capabilities.
Further, the on-device CDKT regularizer helps to reduce the bias and over-fitting issues of local models when training on the small private dataset.
\renewcommand{\arraystretch}{1.06}

\section{Experiments}\label{s:experiment}

\subsection{Setting} 
In this section, we validate the efficacy of the \CDKT algorithm with the 
Fashion-MNIST \citep{xiao2017fashion_mnist}, CIFAR-10, and CIFAR-100 \citep{krizhevsky2009learning} datasets for handwritten digits, fashion images, and object recognition, respectively. We conduct the experiments with $10$ selected clients in each round, where each client has median numbers of data samples at 
$70.5$,  $966.5$, and $1163.5$ with 
Fashion-MNIST, CIFAR-10, and CIFAR-100 respectively. To simulate the properties of non-i.i.d data distribution, we set each client's data from random $20$ classes in $100$ classes for CIFAR-100 dataset and only two classes in total $10$ classes with the other datasets and use $20\%$ of private data samples for evaluating the testing performance. Thereby, the private datasets are unbalanced and have a small number of training samples. \textcolor{black}{We provide the data distributions among clients of different datasets in the Appendix.} Further, the small proxy datasets have all classes data with, respectively, 
$330$, $4200$, and $4200$ samples for 
Fashion-MNIST and CIFAR-10 datasets. They are just $0.5$\% and $7$\% of the overall number of samples and practically accessible by service providers. The global CNN model in the server consists of two convolution layers in 
Fashion-MNIST dataset, whereas three convolution layers are used in the CIFAR-10 and CIFAR-100 dataset, followed by the pooling and fully connected layers \textcolor{black}{(described in Table~\ref{model_architecture_fashion} and Table~\ref{model_architecture_cifar} in Appendix)}. The client CNN models have a similar design or one CNN layer fewer than the global model one in the heterogeneous model setting. The learning parameters, such as learning rates and trade-off parameters, are tuned to achieve good results for different settings of algorithms: the number of local epochs $K = 2$, global epochs $R = 2$, and the batch size is $20$.


\textbf{Evaluation metrics:} To the best of our knowledge, FL approaches focus more on the learning performance of the global model, while personalized FL algorithms evaluate the learning performance of the local models. \textcolor{black}{In this work, to better capture the learning behavior of FL algorithms, we conduct evaluations with the specialization (\textbf{C-Spec}) and generalization (\textbf{C-Gen}) performance of client models on average,  defined by the accuracy and F1 score in their local testing data only, and the collective testing data from all clients, respectively. To summarize the performance of local models, we calculate the average of \textbf{C-Gen} and \textbf{C-Spec} performance as the \textbf{C-Per} metric in terms of accuracy and F1 score.} Using only \textbf{C-Spec} or \textbf{C-Gen} does not fully capture the performance of personalized models. However, considering both metrics allows us to have a complete view of the specialization and the generalization capabilities of personalized models. Therefore, the introduced averaging metric \textbf{C-Per} provides a balance between both capabilities of the personalized models. Meanwhile, the \textbf{Global} metric is the accuracy of the global model for the collective testing data from all clients. 

\textbf{Implementation:} We develop our \CDKT algorithm on Pytorch library \citep{paszke2019pytorch}, which is a robust open-source machine learning framework developed by Facebook. All the experiments are deployed on our server with one NVIDIA GeForce GTX-1080 Ti GPU using CUDA version 11.2 and Intel Core i7-7700K 4.20GHz CPU with sufficient memory for model training. 
\begin{table*}[]
\textcolor{black}{
\caption{\textcolor{black}{The comparison of average accuracy across three datasets over a range of $90$ to $100$ rounds.}}
\label{test_accuracy_100iters}
\begin{center}
\resizebox{14cm}{!}{%
\begin{tabular}{llcccccc}
\hline
\multicolumn{2}{|l|}{\multirow{2}{*}{}}                                                                                                                  & \multicolumn{2}{c|}{\textbf{Fashion-MNIST}}                                & \multicolumn{2}{c|}{\textbf{CIFAR-10}}                                      & \multicolumn{2}{c|}{\textbf{CIFAR-100}}                                    \\ \cline{3-8} 
\multicolumn{2}{|l|}{}                                                                                                                                   & \multicolumn{1}{c|}{\textit{Global}} & \multicolumn{1}{c|}{\textit{C-Per}} & \multicolumn{1}{c|}{\textit{Global}} & \multicolumn{1}{c|}{\textit{C-Pers}} & \multicolumn{1}{c|}{\textit{Global}} & \multicolumn{1}{c|}{\textit{C-Per}} \\ \hline
\multicolumn{1}{|l|}{\multirow{8}{*}{\textbf{\begin{tabular}[c]{@{}l@{}}Fixed\\ Users\end{tabular}}}}     & \multicolumn{1}{l|}{\textbf{No Transfer}}    & \multicolumn{1}{c|}{83.72}           & \multicolumn{1}{c|}{59.09}          & \multicolumn{1}{c|}{55.92}           & \multicolumn{1}{c|}{50.43}           & \multicolumn{1}{c|}{17.27}           & \multicolumn{1}{c|}{18.76}          \\ \cline{2-8} 
\multicolumn{1}{|l|}{}                                                                                    & \multicolumn{1}{l|}{\textbf{FedAvg}}         & \multicolumn{1}{c|}{82.56}           & \multicolumn{1}{c|}{76.41}          & \multicolumn{1}{c|}{59.85}  & \multicolumn{1}{c|}{52.95}           & \multicolumn{1}{c|}{15.53}           & \multicolumn{1}{c|}{17.13}          \\ \cline{2-8} 
\multicolumn{1}{|l|}{}                                                                                    & \multicolumn{1}{l|}{\textbf{Scaffold}}       & \multicolumn{1}{c|}{67.67}           & \multicolumn{1}{c|}{59.59}          & \multicolumn{1}{c|}{48.41}           & \multicolumn{1}{c|}{47.83}           & \multicolumn{1}{c|}{13.48}           & \multicolumn{1}{c|}{13.20}          \\ \cline{2-8} 
\multicolumn{1}{|l|}{}                                                                                    & \multicolumn{1}{l|}{\textbf{MOON}}           & \multicolumn{1}{c|}{83.28}           & \multicolumn{1}{c|}{75.83}          & \multicolumn{1}{c|}{\textbf{60.83}}           & \multicolumn{1}{c|}{53.09}           & \multicolumn{1}{c|}{17.37}           & \multicolumn{1}{c|}{17.67}          \\ \cline{2-8} 
\multicolumn{1}{|l|}{}                                                                                    & \multicolumn{1}{l|}{\textbf{FedDyn}}         & \multicolumn{1}{c|}{83.26}           & \multicolumn{1}{c|}{85.08}          & \multicolumn{1}{c|}{58.00}           & \multicolumn{1}{c|}{59.24}           & \multicolumn{1}{c|}{14.03}           & \multicolumn{1}{c|}{20.23}          \\ \cline{2-8} 
\multicolumn{1}{|l|}{}                                                                                    & \multicolumn{1}{l|}{\textbf{CDKT (Rep)}}     & \multicolumn{1}{c|}{84.88}           & \multicolumn{1}{c|}{\textbf{86.06}} & \multicolumn{1}{c|}{58.44}           & \multicolumn{1}{c|}{\textbf{59.25}}  & \multicolumn{1}{c|}{17.75}           & \multicolumn{1}{c|}{21.81}          \\ \cline{2-8} 
\multicolumn{1}{|l|}{}                                                                                    & \multicolumn{1}{l|}{\textbf{CDKT (Full)}}    & \multicolumn{1}{c|}{85.81}           & \multicolumn{1}{c|}{81.66}          & \multicolumn{1}{c|}{58.91}           & \multicolumn{1}{c|}{55.83}           & \multicolumn{1}{c|}{17.84}           & \multicolumn{1}{c|}{21.93}          \\ \cline{2-8} 
\multicolumn{1}{|l|}{}                                                                                    & \multicolumn{1}{l|}{\textbf{CDKT (RepFull)}} & \multicolumn{1}{c|}{\textbf{86.51}}  & \multicolumn{1}{c|}{84.08}          & \multicolumn{1}{c|}{59.09}           & \multicolumn{1}{c|}{57.74}           & \multicolumn{1}{c|}{\textbf{18.49}}  & \multicolumn{1}{c|}{\textbf{22.22}} \\ \hline
                                                                                                          &                                              &                                      &                                     &                                      &                                      &                                      &                                     \\ \hline
\multicolumn{1}{|l|}{\multirow{8}{*}{\textbf{\begin{tabular}[c]{@{}l@{}}Subset\\ of Users\end{tabular}}}} & \multicolumn{1}{l|}{\textbf{No Transfer}}    & \multicolumn{1}{c|}{78.19}           & \multicolumn{1}{c|}{58.98}          & \multicolumn{1}{c|}{52.17}           & \multicolumn{1}{c|}{46.75}           & \multicolumn{1}{c|}{16.47}           & \multicolumn{1}{c|}{8.13}           \\ \cline{2-8} 
\multicolumn{1}{|l|}{}                                                                                    & \multicolumn{1}{l|}{\textbf{FedAvg}}         & \multicolumn{1}{c|}{69.44}           & \multicolumn{1}{c|}{70.62}          & \multicolumn{1}{c|}{48.83}           & \multicolumn{1}{c|}{47.20}           & \multicolumn{1}{c|}{12.25}           & \multicolumn{1}{c|}{12.57}          \\ \cline{2-8} 
\multicolumn{1}{|l|}{}                                                                                    & \multicolumn{1}{l|}{\textbf{Scaffold}}       & \multicolumn{1}{c|}{68.00}           & \multicolumn{1}{c|}{67.88}          & \multicolumn{1}{c|}{40.76}           & \multicolumn{1}{c|}{43.06}           & \multicolumn{1}{c|}{12.29}           & \multicolumn{1}{c|}{12.12}          \\ \cline{2-8} 
\multicolumn{1}{|l|}{}                                                                                    & \multicolumn{1}{l|}{\textbf{MOON}}           & \multicolumn{1}{c|}{70.05}           & \multicolumn{1}{c|}{70.70}          & \multicolumn{1}{c|}{53.05}           & \multicolumn{1}{c|}{52.15}           & \multicolumn{1}{c|}{16.79}           & \multicolumn{1}{c|}{16.69}          \\ \cline{2-8} 
\multicolumn{1}{|l|}{}                                                                                    & \multicolumn{1}{l|}{\textbf{FedDyn}}         & \multicolumn{1}{c|}{76.03}           & \multicolumn{1}{c|}{76.11}          & \multicolumn{1}{c|}{55.71}           & \multicolumn{1}{c|}{51.91}           & \multicolumn{1}{c|}{16.37}           & \multicolumn{1}{c|}{16.49}          \\ \cline{2-8} 
\multicolumn{1}{|l|}{}                                                                                    & \multicolumn{1}{l|}{\textbf{CDKT (Rep)}}     & \multicolumn{1}{c|}{78.50}           & \multicolumn{1}{c|}{80.90}          & \multicolumn{1}{c|}{55.67}           & \multicolumn{1}{c|}{53.92}           & \multicolumn{1}{c|}{16.72}           & \multicolumn{1}{c|}{15.74}          \\ \cline{2-8} 
\multicolumn{1}{|l|}{}                                                                                    & \multicolumn{1}{l|}{\textbf{CDKT (Full)}}    & \multicolumn{1}{c|}{79.22}           & \multicolumn{1}{c|}{79.02}          & \multicolumn{1}{c|}{\textbf{56.60}}  & \multicolumn{1}{c|}{\textbf{54.35}}  & \multicolumn{1}{c|}{17.33}           & \multicolumn{1}{c|}{15.76}          \\ \cline{2-8} 
\multicolumn{1}{|l|}{}                                                                                    & \multicolumn{1}{l|}{\textbf{CDKT (RepFull)}} & \multicolumn{1}{c|}{\textbf{79.94}}  & \multicolumn{1}{c|}{\textbf{81.39}} & \multicolumn{1}{c|}{56.57}           & \multicolumn{1}{c|}{54.03}           & \multicolumn{1}{c|}{\textbf{17.60}}  & \multicolumn{1}{c|}{\textbf{17.74}} \\ \hline
\end{tabular}
}%
\end{center}
}
\end{table*}
\begin{table}[t]
\textcolor{black}{
\caption{\textcolor{black}{The comparison of average F1 score (weighted) across three datasets over a range of $90$ to $100$ rounds.}}
\label{F1_score_100iters}
\begin{center}
\resizebox{14cm}{!}{%
\begin{tabular}{llcccccc}
\hline
\multicolumn{2}{|l|}{\multirow{2}{*}{}}                                                                                                                  & \multicolumn{2}{c|}{\textbf{Fashion-MNIST}}                                & \multicolumn{2}{c|}{\textbf{CIFAR-10}}                                      & \multicolumn{2}{c|}{\textbf{CIFAR-100}}                                    \\ \cline{3-8} 
\multicolumn{2}{|l|}{}                                                                                                                                   & \multicolumn{1}{c|}{\textit{Global}} & \multicolumn{1}{c|}{\textit{C-Per}} & \multicolumn{1}{c|}{\textit{Global}} & \multicolumn{1}{c|}{\textit{C-Pers}} & \multicolumn{1}{c|}{\textit{Global}} & \multicolumn{1}{c|}{\textit{C-Per}} \\ \hline
\multicolumn{1}{|l|}{\multirow{8}{*}{\textbf{\begin{tabular}[c]{@{}l@{}}Fixed\\ Users\end{tabular}}}}     & \multicolumn{1}{l|}{\textbf{No Transfer}}    & \multicolumn{1}{c|}{86.65}           & \multicolumn{1}{c|}{58.82}          & \multicolumn{1}{c|}{64.17}           & \multicolumn{1}{c|}{48.24}           & \multicolumn{1}{c|}{20.68}           & \multicolumn{1}{c|}{14.44}          \\ \cline{2-8} 
\multicolumn{1}{|l|}{}                                                                                    & \multicolumn{1}{l|}{\textbf{FedAvg}}         & \multicolumn{1}{c|}{81.75}           & \multicolumn{1}{c|}{75.53}          & \multicolumn{1}{c|}{66.46}           & \multicolumn{1}{c|}{51.85}           & \multicolumn{1}{c|}{18.19}           & \multicolumn{1}{c|}{12.79}          \\ \cline{2-8} 
\multicolumn{1}{|l|}{}                                                                                    & \multicolumn{1}{l|}{\textbf{Scaffold}}       & \multicolumn{1}{c|}{77.73}           & \multicolumn{1}{c|}{58.92}          & \multicolumn{1}{c|}{57.52}           & \multicolumn{1}{c|}{46.14}           & \multicolumn{1}{c|}{15.40}           & \multicolumn{1}{c|}{10.19}          \\ \cline{2-8} 
\multicolumn{1}{|l|}{}                                                                                    & \multicolumn{1}{l|}{\textbf{MOON}}           & \multicolumn{1}{c|}{85.53}           & \multicolumn{1}{c|}{79.96}          & \multicolumn{1}{c|}{\textbf{69.03}}  & \multicolumn{1}{c|}{56.79}           & \multicolumn{1}{c|}{20.81}           & \multicolumn{1}{c|}{12.97}          \\ \cline{2-8} 
\multicolumn{1}{|l|}{}                                                                                    & \multicolumn{1}{l|}{\textbf{FedDyn}}         & \multicolumn{1}{c|}{85.11}           & \multicolumn{1}{c|}{85.97}          & \multicolumn{1}{c|}{65.93}           & \multicolumn{1}{c|}{62.34}           & \multicolumn{1}{c|}{19.72}           & \multicolumn{1}{c|}{21.54}          \\ \cline{2-8} 
\multicolumn{1}{|l|}{}                                                                                    & \multicolumn{1}{l|}{\textbf{CDKT (Rep)}}     & \multicolumn{1}{c|}{\textbf{87.04}}  & \multicolumn{1}{c|}{\textbf{89.85}} & \multicolumn{1}{c|}{67.15}           & \multicolumn{1}{c|}{\textbf{62.40}}  & \multicolumn{1}{c|}{23.60}           & \multicolumn{1}{c|}{\textbf{21.67}} \\ \cline{2-8} 
\multicolumn{1}{|l|}{}                                                                                    & \multicolumn{1}{l|}{\textbf{CDKT (Full)}}    & \multicolumn{1}{c|}{86.37}           & \multicolumn{1}{c|}{84.52}          & \multicolumn{1}{c|}{67.21}           & \multicolumn{1}{c|}{58.32}           & \multicolumn{1}{c|}{23.37}           & \multicolumn{1}{c|}{19.65}          \\ \cline{2-8} 
\multicolumn{1}{|l|}{}                                                                                    & \multicolumn{1}{l|}{\textbf{CDKT (RepFull)}} & \multicolumn{1}{c|}{86.69}           & \multicolumn{1}{c|}{82.82}          & \multicolumn{1}{c|}{68.15}           & \multicolumn{1}{c|}{60.16}           & \multicolumn{1}{c|}{\textbf{24.68}}  & \multicolumn{1}{c|}{21.30}          \\ \hline
                                                                                                          &                                              &                                      &                                     &                                      &                                      &                                      &                                     \\ \hline
\multicolumn{1}{|l|}{\multirow{8}{*}{\textbf{\begin{tabular}[c]{@{}l@{}}Subset\\ of Users\end{tabular}}}} & \multicolumn{1}{l|}{\textbf{No Transfer}}    & \multicolumn{1}{c|}{85.65}           & \multicolumn{1}{c|}{57.92}          & \multicolumn{1}{c|}{64.84}           & \multicolumn{1}{c|}{35.97}           & \multicolumn{1}{c|}{21.57}           & \multicolumn{1}{c|}{4.98}           \\ \cline{2-8} 
\multicolumn{1}{|l|}{}                                                                                    & \multicolumn{1}{l|}{\textbf{FedAvg}}         & \multicolumn{1}{c|}{73.91}           & \multicolumn{1}{c|}{75.65}          & \multicolumn{1}{c|}{61.54}           & \multicolumn{1}{c|}{58.48}           & \multicolumn{1}{c|}{12.50}           & \multicolumn{1}{c|}{12.33}          \\ \cline{2-8} 
\multicolumn{1}{|l|}{}                                                                                    & \multicolumn{1}{l|}{\textbf{Scaffold}}       & \multicolumn{1}{c|}{75.51}           & \multicolumn{1}{c|}{83.19}          & \multicolumn{1}{c|}{51.34}           & \multicolumn{1}{c|}{50.37}           & \multicolumn{1}{c|}{14.04}           & \multicolumn{1}{c|}{13.59}          \\ \cline{2-8} 
\multicolumn{1}{|l|}{}                                                                                    & \multicolumn{1}{l|}{\textbf{MOON}}           & \multicolumn{1}{c|}{74.53}           & \multicolumn{1}{c|}{76.90}          & \multicolumn{1}{c|}{65.00}           & \multicolumn{1}{c|}{62.72}           & \multicolumn{1}{c|}{18.85}           & \multicolumn{1}{c|}{13.84}          \\ \cline{2-8} 
\multicolumn{1}{|l|}{}                                                                                    & \multicolumn{1}{l|}{\textbf{FedDyn}}         & \multicolumn{1}{c|}{81.33}           & \multicolumn{1}{c|}{81.55}          & \multicolumn{1}{c|}{64.32}           & \multicolumn{1}{c|}{60.64}           & \multicolumn{1}{c|}{18.54}           & \multicolumn{1}{c|}{14.23}          \\ \cline{2-8} 
\multicolumn{1}{|l|}{}                                                                                    & \multicolumn{1}{l|}{\textbf{CDKT (Rep)}}     & \multicolumn{1}{c|}{86.41}           & \multicolumn{1}{c|}{\textbf{85.96}} & \multicolumn{1}{c|}{65.69}           & \multicolumn{1}{c|}{62.41}           & \multicolumn{1}{c|}{22.61}           & \multicolumn{1}{c|}{\textbf{14.45}} \\ \cline{2-8} 
\multicolumn{1}{|l|}{}                                                                                    & \multicolumn{1}{l|}{\textbf{CDKT (Full)}}    & \multicolumn{1}{c|}{86.39}           & \multicolumn{1}{c|}{83.35}          & \multicolumn{1}{c|}{\textbf{67.07}}  & \multicolumn{1}{c|}{60.68}  & \multicolumn{1}{c|}{22.86}           & \multicolumn{1}{c|}{12.94}          \\ \cline{2-8} 
\multicolumn{1}{|l|}{}                                                                                    & \multicolumn{1}{l|}{\textbf{CDKT (RepFull)}} & \multicolumn{1}{c|}{\textbf{86.46}}  & \multicolumn{1}{c|}{85.90}          & \multicolumn{1}{c|}{66.63}           & \multicolumn{1}{c|}{\textbf{63.01}}           & \multicolumn{1}{c|}{\textbf{22.67}}  & \multicolumn{1}{c|}{13.10}          \\ \hline
\end{tabular}
}%
\end{center}
}
\end{table}

\textbf{Baselines and CDKT settings:} \textcolor{black}{\textcolor{black}{For comparison, we evaluate different settings of the \CDKT algorithm and compared them with serveral state-of-the-art approaches including (1) FedAvg~\citep{mcmahan2016communication}, (2) SCAFFOLD~\citep{karimireddy2020scaffold}, (3) MOON~\citep{li2021model} and (4) FedDyn~\citep{acar2021federated}.} In \textbf{No Transfer} setting, the clients use the private DS, and the server uses the proxy DS. Accordingly, the setting \textbf{Rep}, \textbf{Full}, and \textbf{RepFull} stand for using only the embedding features of representation, outcomes of full models, and both in the global knowledge transfer, respectively.} 
Furthermore, different distance functions were also validated, such as KL, JS, and Norm-2 (N). The setting KL-N indicates the use of KL divergence in the global CDKT regularizer, while the on-device CDKT regularizer uses Norm-2.



\begin{figure*}[]
	\centering
	\includegraphics[width=\linewidth]{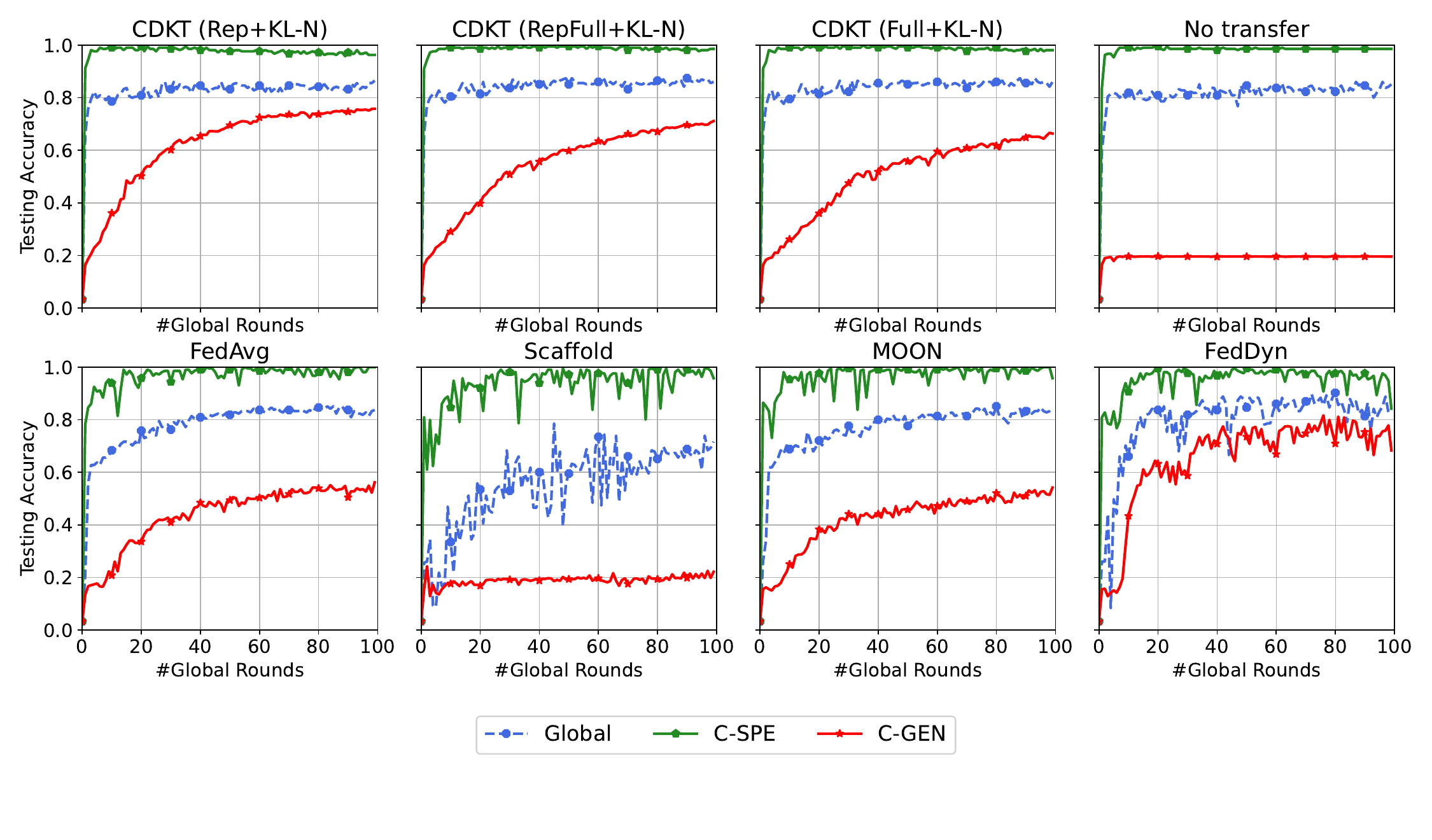}
	\caption{\textcolor{black}{Comparative Analysis of \CDKT under various settings against other baselines using the Fashion-MNIST dataset in Fixed Users scenario.}}
	\label{fashionmnist_fixed}
\end{figure*}

\begin{figure*}[]
	\centering
	\includegraphics[width=\linewidth]{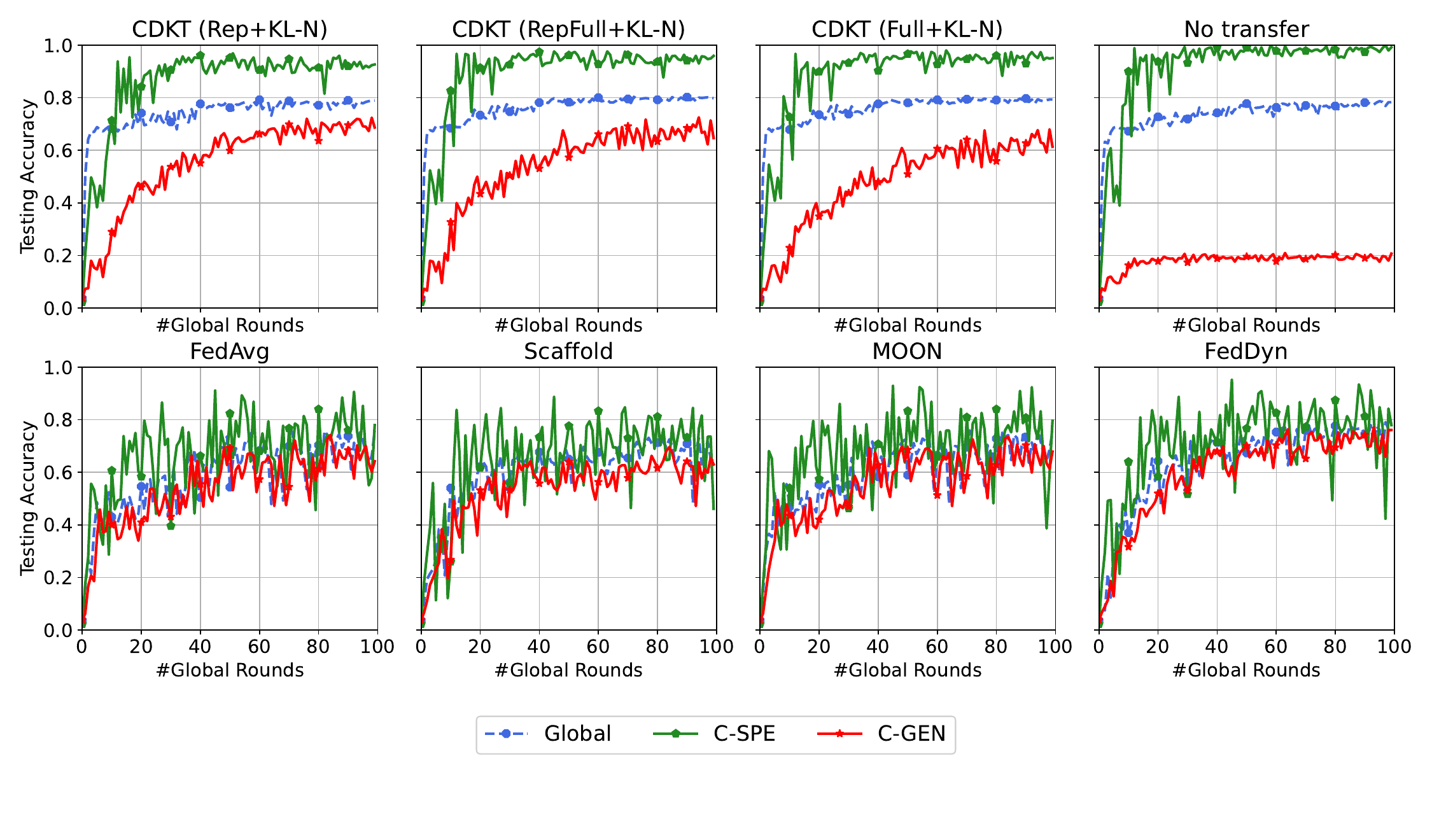}
	\caption{\textcolor{black}{Comparative Analysis of \CDKT under various settings against other baselines using the Fashion-MNIST dataset in Subset of Users scenario.}}
	\label{fashionmnist_subset}
\end{figure*}

\begin{figure*}[t]
 \centering
	\includegraphics[width=1.\linewidth]{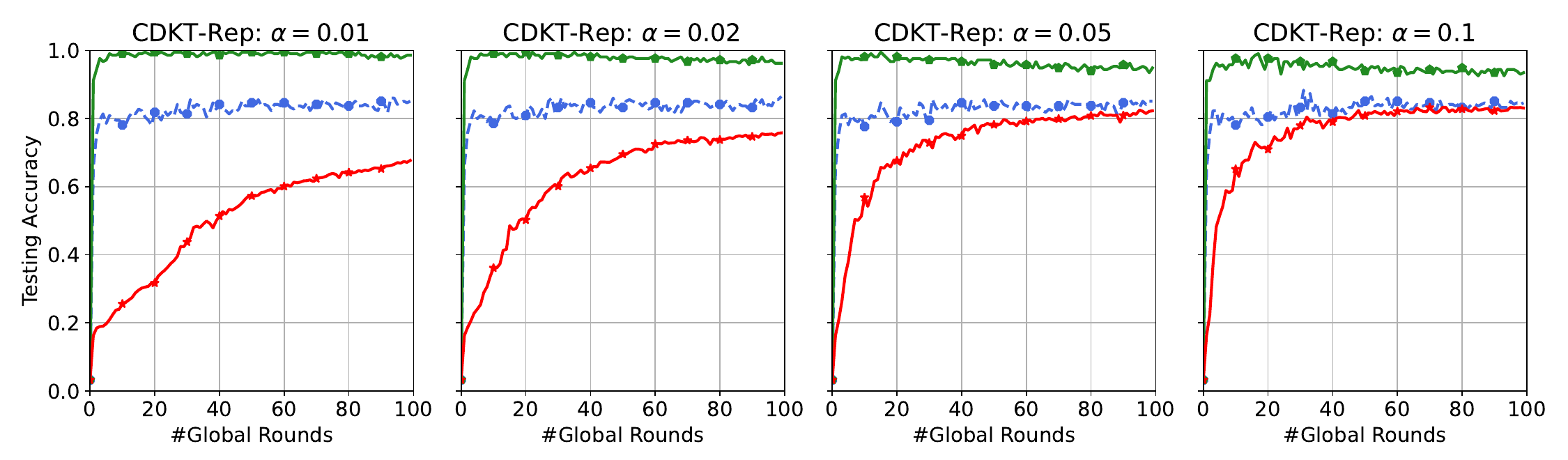}
	\label{fmnist_alpha}
	\includegraphics[width=1.\linewidth]{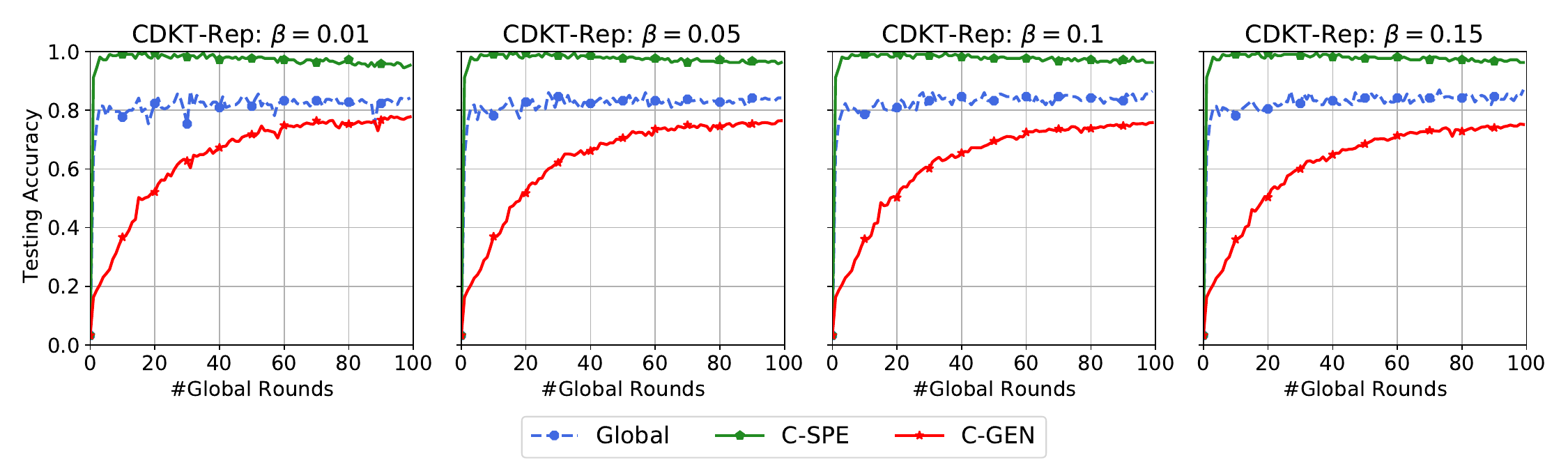}
	\label{fmnist_beta}
	\caption{Effects of trade-off parameters in \CDKT with Fashion-MNIST dataset in Fixed Users scenario.}
	\label{Fig:parameter_effect_fmnist2}
\end{figure*}

\begin{figure*}[t]
 \centering
	\includegraphics[width=1.\linewidth]{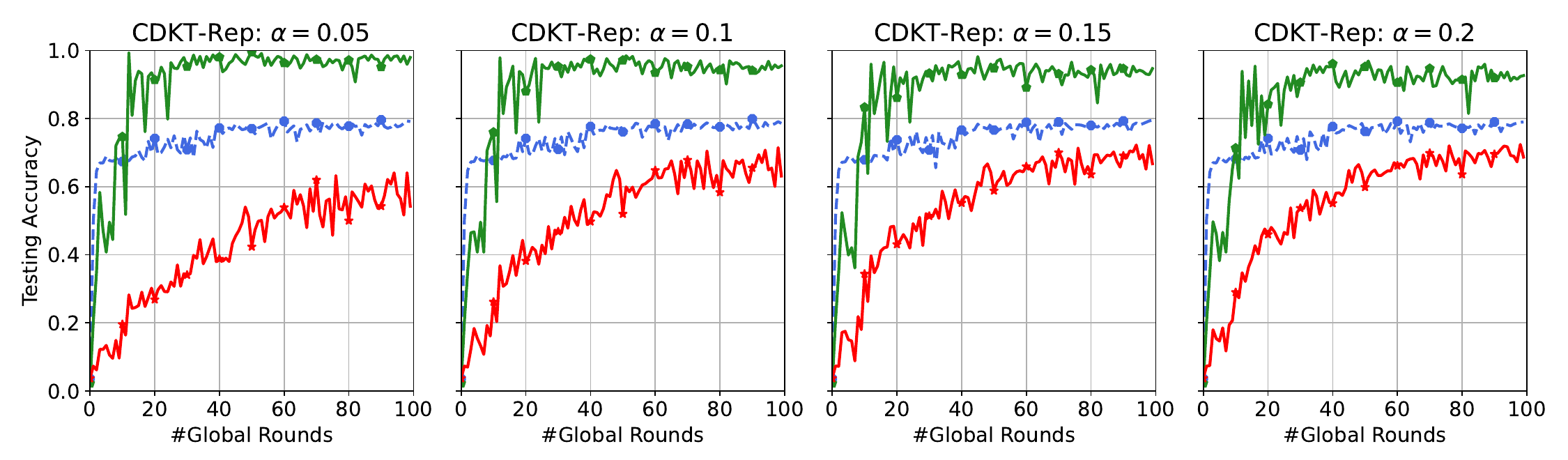}
	\label{fmnist_alpha}
	\includegraphics[width=1.\linewidth]{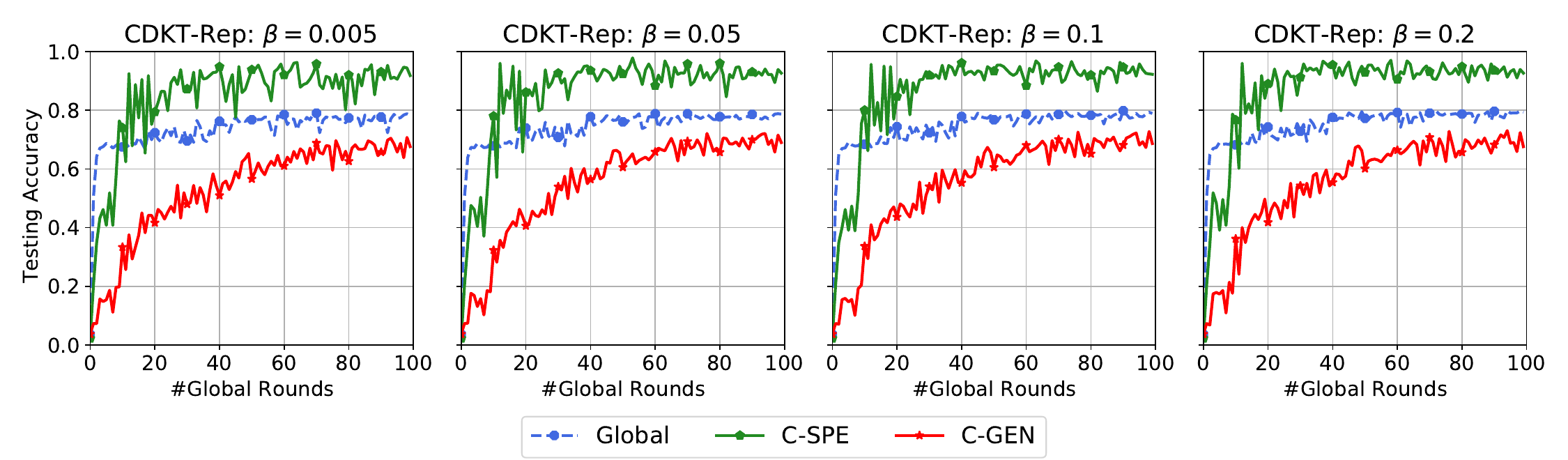}
	\label{fmnist_beta}
	\caption{Effects of trade-off parameters in \CDKT with Fashion-MNIST dataset in Subset of Users scenario.}
	\label{Fig:parameter_effect_fmnist1}
\end{figure*}
\subsection{Experimental Results}

\begin{figure*}[]
 \centering
    \begin{subfigure}{\linewidth}
	\includegraphics[width=\linewidth]{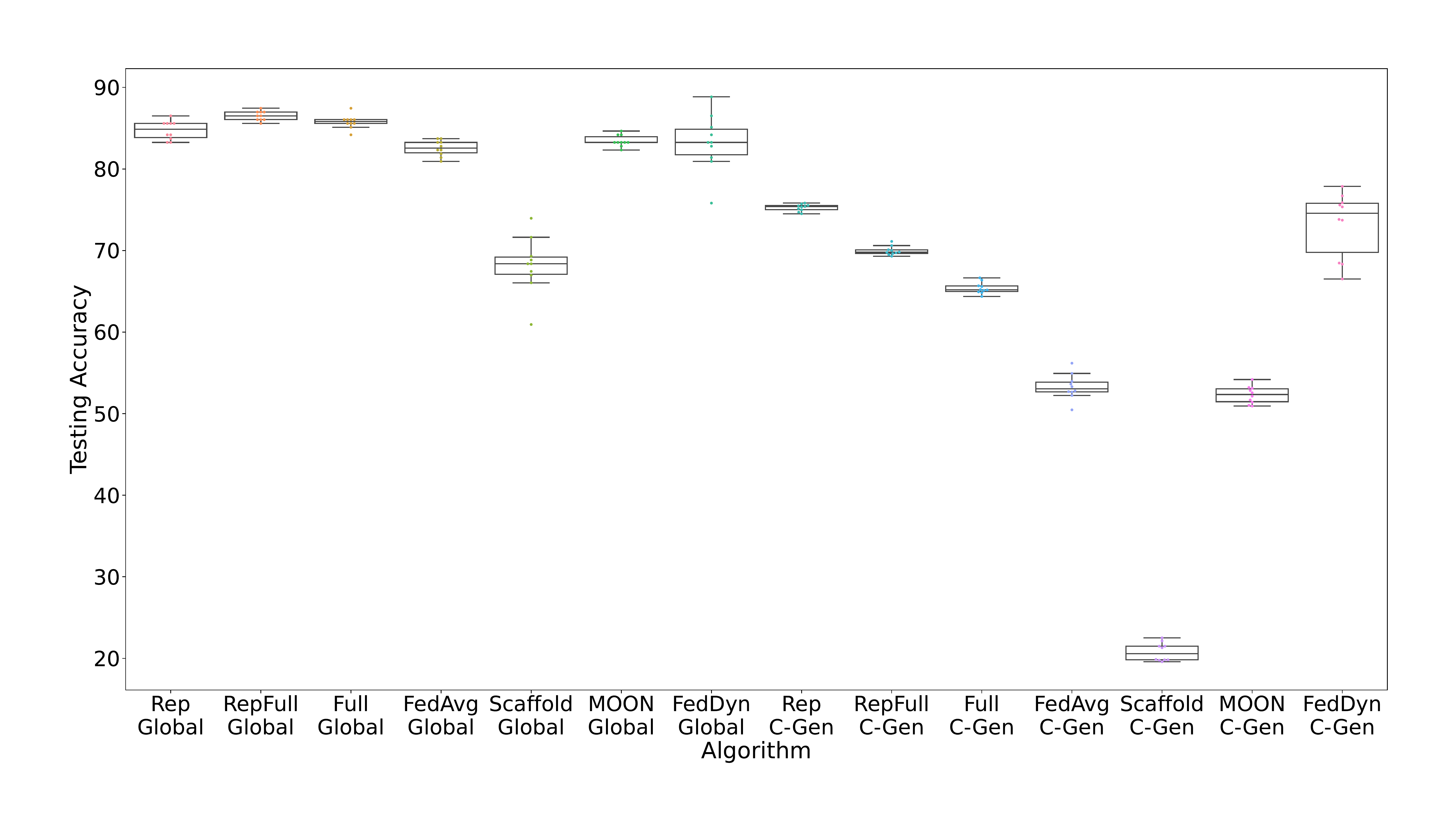}
	\caption{Fixed Users Scenario}
	\label{fmnist_fixed_box}
    \end{subfigure}
    \begin{subfigure}{\linewidth}
	\includegraphics[width=\linewidth]{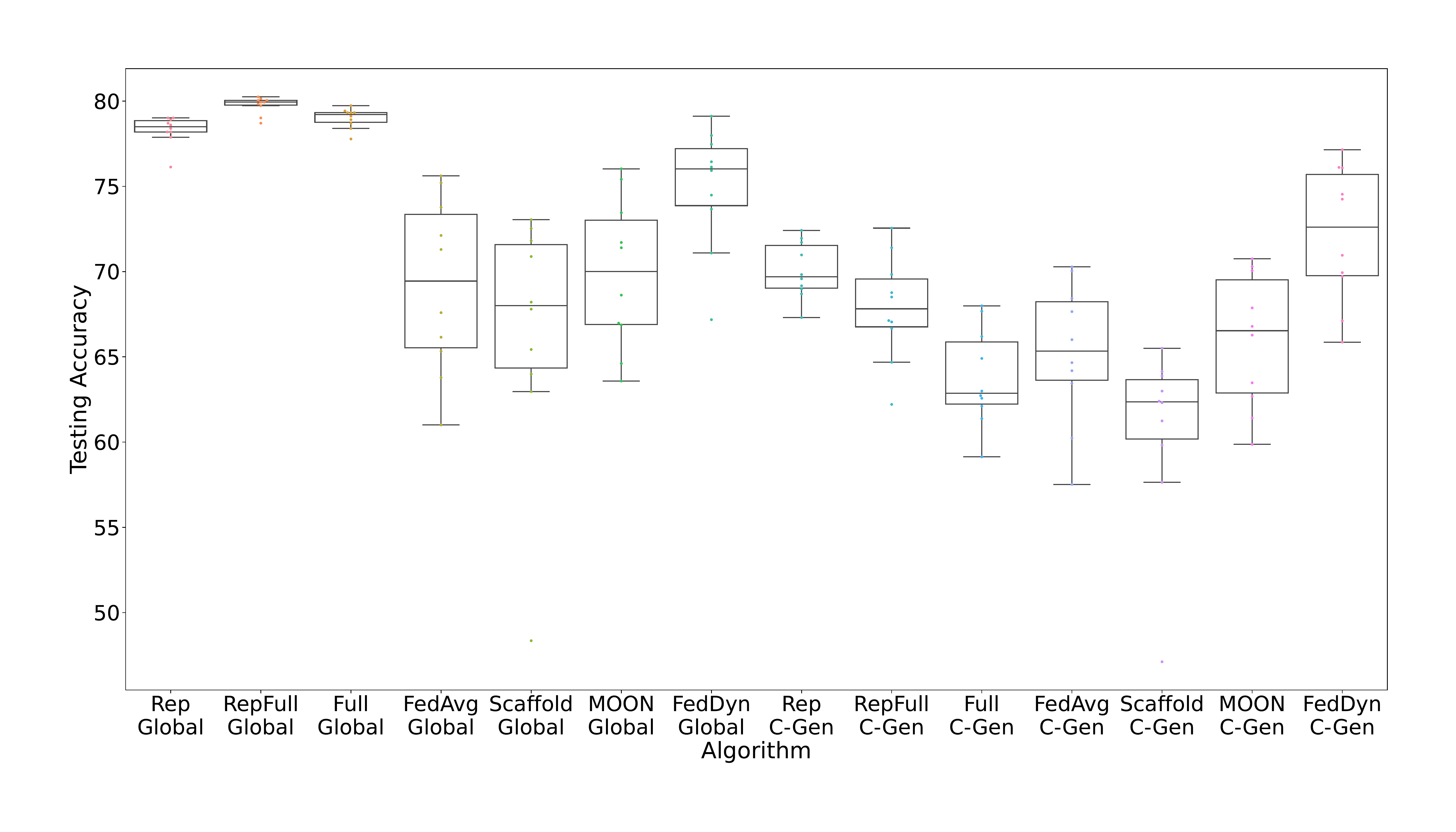}
	\caption{Subset of Users Scenario}
	\label{fmnist_subset_box}
	\end{subfigure}
	\caption{\textcolor{black}{Convergence analysis of \CDKT under various settings against other baselines using the Fashion-MNIST dataset.}}
	\label{fmnist_box}
\end{figure*}

\begin{figure*}[]
 \centering
    \begin{subfigure}{\linewidth}
	\includegraphics[width=\linewidth]{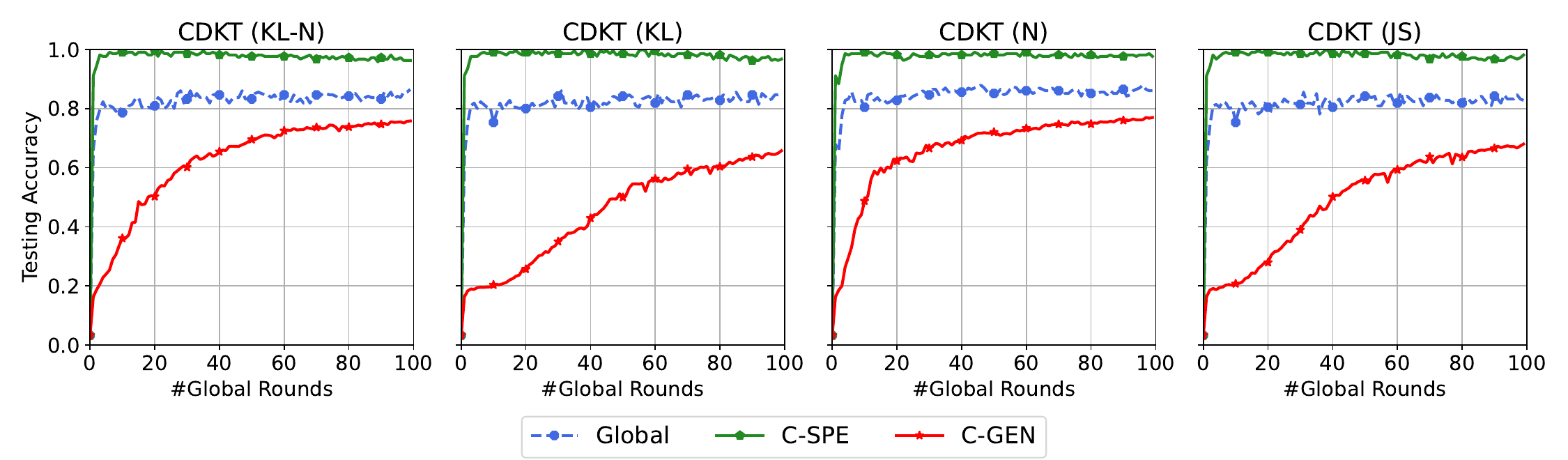}
	\caption{Fixed Users Scenario}
	\label{fmnist_metric_fixed}
	\end{subfigure}
    \begin{subfigure}{\linewidth}
	\includegraphics[width=\linewidth]{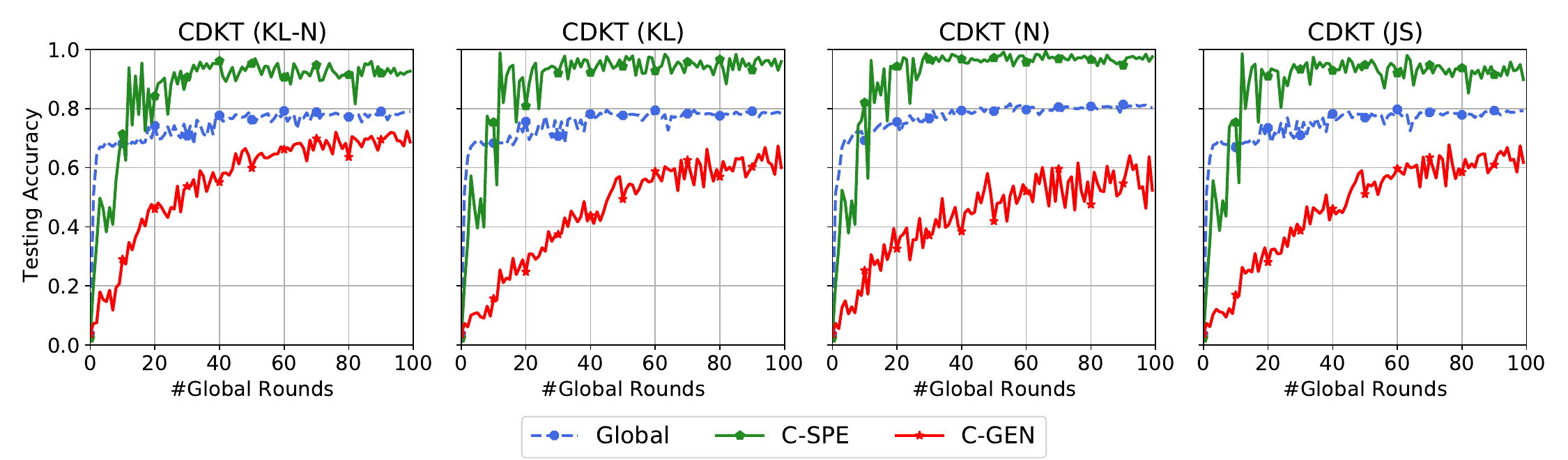}
	\caption{Subset of Users Scenario}
	\label{fmnist_metric_subset}
	\end{subfigure}
    \caption{Performance analysis of \CDKT with different distance metrics on the Fashion-MNIST dataset.}
	\label{fmnist_metric}
\end{figure*}

\begin{figure*}
    \centering
    \begin{subfigure}{0.47\textwidth}
        \centering
        \includegraphics[width=1.1\linewidth]{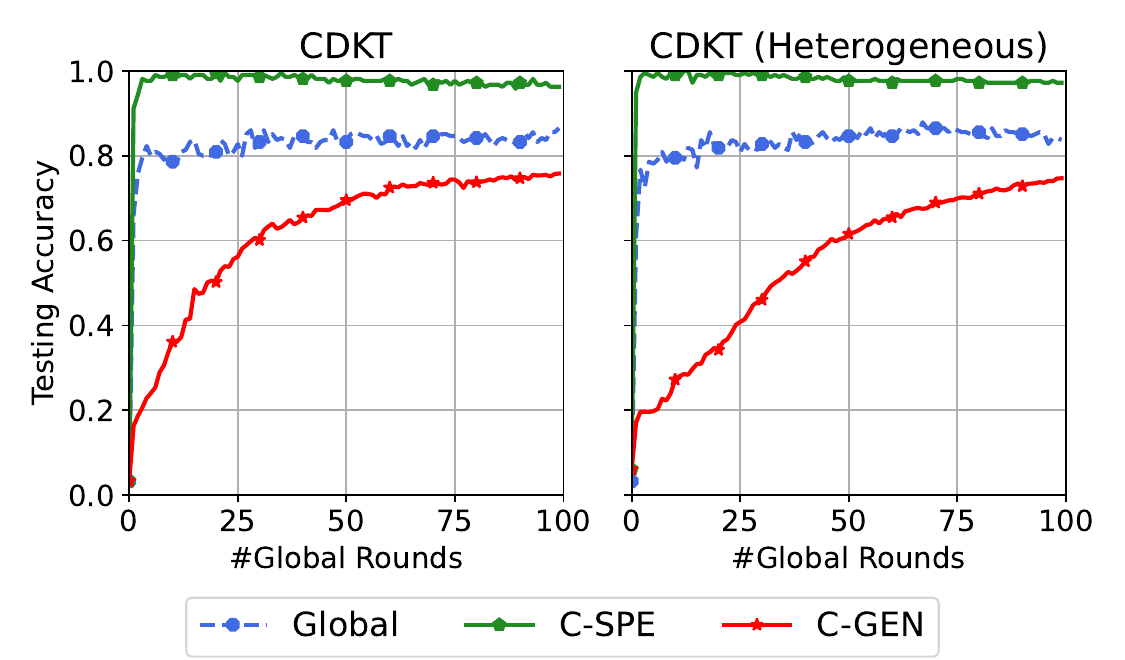}
        \caption{Fixed Users Scenario}
        \label{fmnist_metric_fixed}
    \end{subfigure}\hfill
    \begin{subfigure}{0.47\textwidth}
        \centering
        \includegraphics[width=1.1\linewidth]{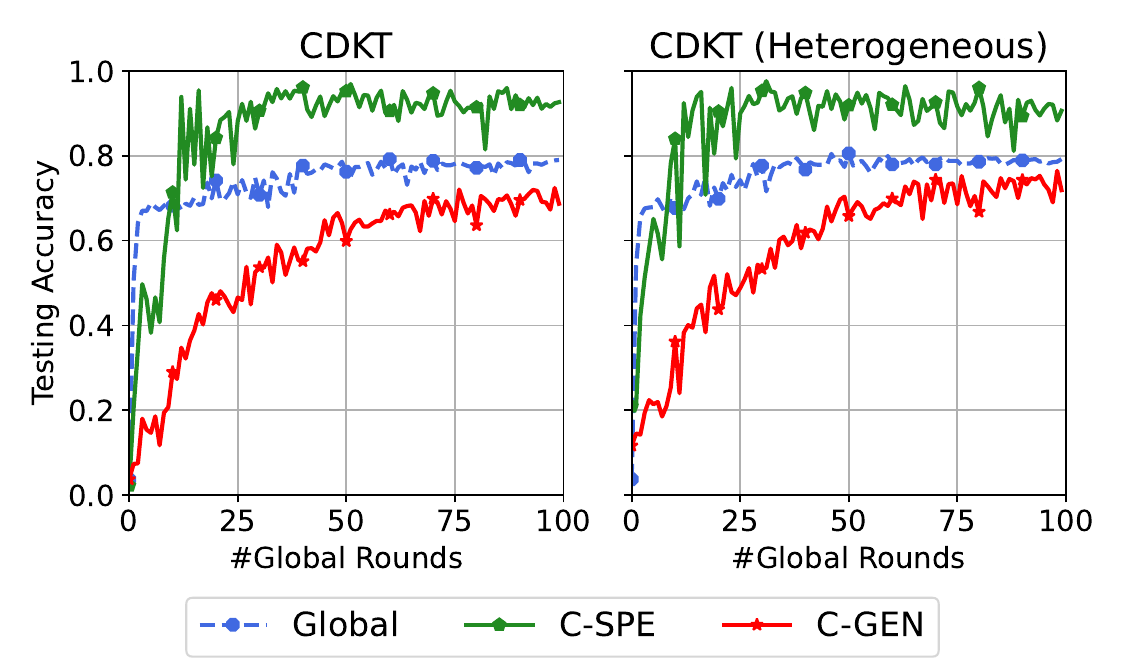}
        \caption{Subset of Users Scenario}
        \label{fmnist_metric_subset}
    \end{subfigure}
    \caption{Performance analysis of \CDKT with heterogeneous models on the Fashion-MNIST dataset.}
    \label{fmnist_same_hetero}
\end{figure*}
\textcolor{black}{In summary, we obtained experimental results on three datasets for different settings in the \CDKT algorithm and baselines with the median testing accuracy and F1 score from round $90$ to $100$, as in Table~\ref{test_accuracy_100iters} and Table~\ref{F1_score_100iters}. There are two principal scenarios: 1) the Fixed Users scenario, where all $10$ users participate in every communication round, and (2) the Subset of Users scenario, in which 10 randomly selected users from a total of $50$ users in each communication round. Compared to the other baselines, in terms of accuracy metric, \CDKT outperforms $10\%$ and $5\%$ with Fashion-MNIST, $6\%$ and $2\%$  with CIFAR-10, $2\%$ and $1\%$ with CIFAR-100 for scenario (1) and scenario (2) respectively, according to \textbf{C-Per} performance. In addition to the main target to improve personalized performance, in both metrics, \CDKT also significantly improves the performance of the global model in most of the settings as shown in Table~\ref{test_accuracy_100iters} and Table~\ref{F1_score_100iters}. According to our detailed figures, in scenario (1), \CDKT obtains a higher \textbf{C-Gen} performance, a comparable \textbf{C-Spec} performance of client models, and a comparable or higher \textbf{Global} performance in all three datasets. In scenario (2), our proposed algorithm achieves higher \textbf{C-Spec}, \textbf{Global} performance and comparable \textbf{C-Gen} performance. We provide the detailed results of \textbf{C-Spec} and \textbf{C-Gen} in the Appendix. In principle, CDKT helps the client models improve \textbf{C-Gen} and slightly improve the \textbf{Global} performance than that of no transfer setting. In addition, as the final performance metric, the \CDKT algorithm exhibits better stability and faster convergence while improving the \textbf{C-Per}. It preserves high \textbf{Global} performance in most settings. These attributes also enable the client or global models to be usable early after $50$ global rounds. As expected, in the subset of users selection scenario, other baseline show an extensive fluctuation due to irregularity in learning parameters across the client models stemming from statistical heterogeneity.}


\textcolor{black}{In Fig.~\ref{fashionmnist_fixed} and Fig.~\ref{fashionmnist_subset}, we evaluate different settings knowledge transfer schemes of \CDKT to compare with other baselines in both of the scenarios on Fashion-MNIST dataset. In scenario (1), as shown in Fig.~\ref{fashionmnist_fixed}, experimental results demonstrate that our proposed schemes outperform other baselines in terms of speedups \textbf{C-Gen} and \textbf{Global} performance. The \textbf{Rep+KL-N} setting achieves the highest \textbf{C-Gen} performance with $75.37\%$ but slightly lower \textbf{C-Spec} than that of MOON. In scenario (2), Fig.~\ref{fashionmnist_subset} shows that the \CDKT algorithms with three settings significantly improve \textbf{C-Spec}, \textbf{Global} and more stable than other approaches. We also provide the figures for CIFAR-10 and CIFAR-100 in the Appendix. }

We next evaluate the effects of the trade-off parameters $\alpha$ and $\beta$ using to control the compensation between learning loss and the CDKT regularizer as shown in Fig. \ref{Fig:parameter_effect_fmnist2} and Fig.~\ref{Fig:parameter_effect_fmnist1} in both scenarios, respectively. 
As a result, when increasing $\alpha$, we can boost \textbf{C-Gen} performance; however, that slightly degrades the \textbf{C-Spec} performance. Such learning behavior shows the interesting compensation effect between \textbf{C-Gen} and \textbf{C-Spec} performances, such as till some threshold the specialized capability may decrease as increasing the generalized capability of client models. On the other hand, $\beta$ does not show significant effects on the performance, however, when $\beta$ is too small, it makes the global performance is fluctuated due to less contribution from CDKT regularizers and transferred knowledge. Finally, we evaluate the effectiveness of the \CDKT algorithm using heterogeneous local models, i.e., using one CNN layer less than global model. The \CDKT algorithm with smaller client models achieves comparable performance in Fashion-MNIST dataset but slight degradation in the other datasets. 

\textcolor{black}{Additionally, In Fig.~\ref{fmnist_box}, we plot the box plot for different settings of \CDKT to compare with other baselines in terms of stability in the last 20 rounds in both scenarios. As can be seen, our algorithm obtains a very stable convergence with better performance while other methods depict the extensive fluctuation, especially in the subset of users scenario. For Fashion-MNIST, we also evaluate the effect of different distance metrics in CDKT-regularizers in both scenarios in Fig.~\ref{fmnist_metric}. We observe that the setting KL-N obtains stable and higher performance among other metrics. In the KL-N setting, KL divergence is used in the global CDKT regularizer, while the on-device CDKT regularizer uses Norm. Finally, we observe that our method obtains the comparable performance when using heterogeneous models in Fig.~\ref{fmnist_same_hetero} with just a slight degradation in fixed users scenario.}

\section{Conclusion}\label{s:conclusion}
This work bridges the gap between existing alignment methods and knowledge distillation to develop a general form of cross-device knowledge transfer in FL. \textcolor{black}{Through extensive experiments on numerous scenarios, the performance of client and global models are validated for heterogeneous FL. The detailed results following extensive simulations on three datasets have shown the advantages of the proposed algorithms.} Compared to the other methods, we demonstrate that the \CDKT algorithm is more stable and can speed up the global model performance and higher personalized performance of client models. In addition to the learning performance, the proposed algorithm enables (i) the flexibility of heterogeneous model architectures, (ii) reduces communication data with the outcome, and (or) representation knowledge instead of full model parameters compared to the conventional FL algorithms, and (iii) tackles privacy leakage issues (via reverse engineering techniques) of current FL algorithms. As a trade-off, the proposed algorithm requires an additional small proxy dataset for transferring knowledge.
Finally, we advocate that the knowledge transfer method works as a hybrid design with recent advanced techniques to improve learning performance.
\label{}

\bibliographystyle{elsarticle-harv} 
\bibliography{Dem-AI, CDKT}

\appendix
\section{Supplemental Results}
\subsection{Cross Entropy Loss:}
$${\ell}_{CE}(z|D_r)=  -\sum_{ x \in D_r}\sum_{c}P(y_{c}|x) \ln P(z_{c}|x),$$
 where $P(z_{c}|x)$ is the prediction outcome probability and $P(y_{c}|x)$ is the target label of class $c$ given data sample $x$.
 \subsection{CIFAR-10 and CIFAR-100 results}
\begin{figure*}[]
	\centering
	\includegraphics[width=\linewidth]{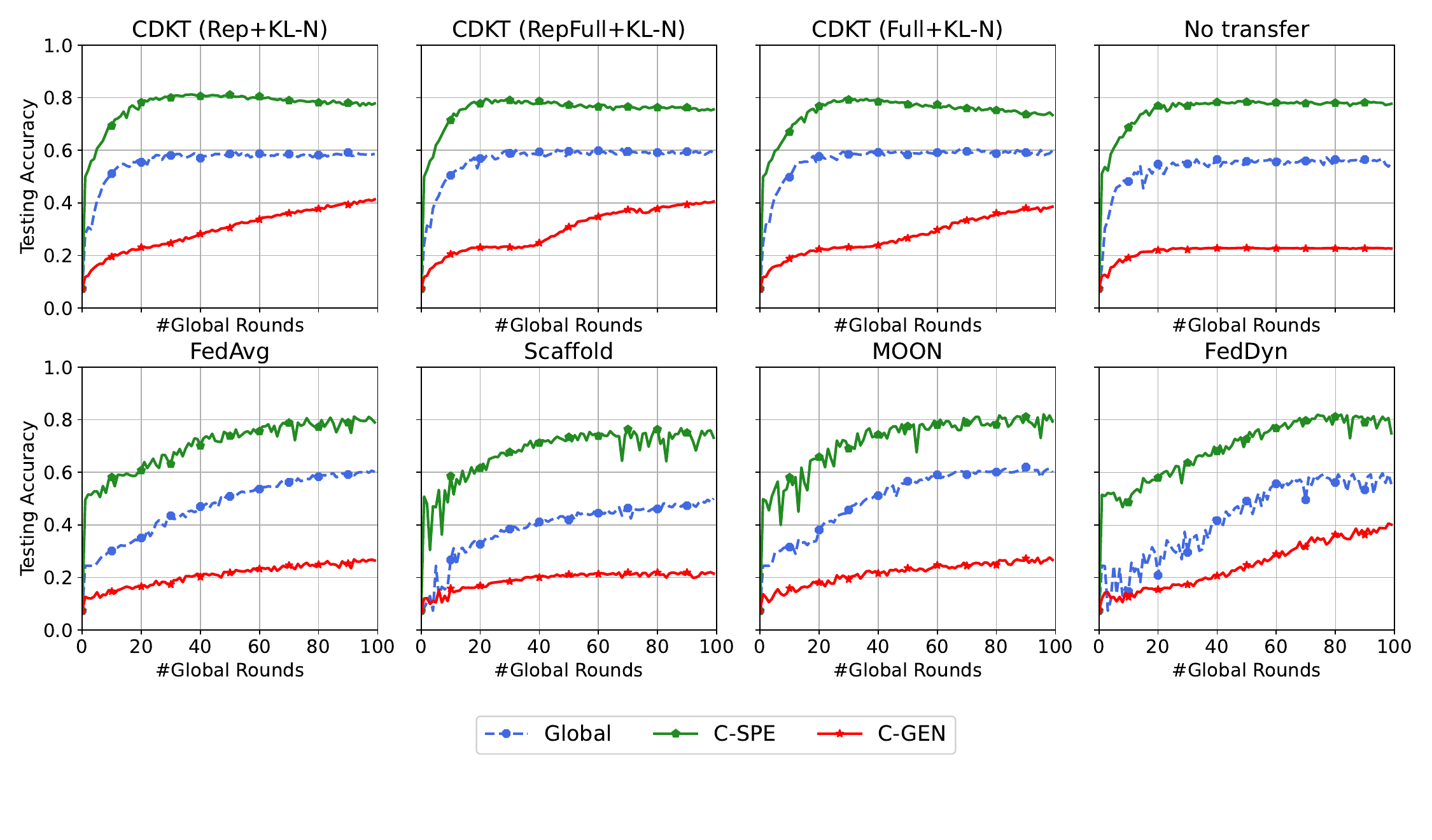}
	\caption{\textcolor{black}{Comparative analysis of \CDKT under various settings against other baselines using the CIFAR-10 dataset in Fixed Users scenario.}}
	\label{fixed_users_cifar10}
\end{figure*}

\begin{figure*}[]
	\centering
	\includegraphics[width=\linewidth]{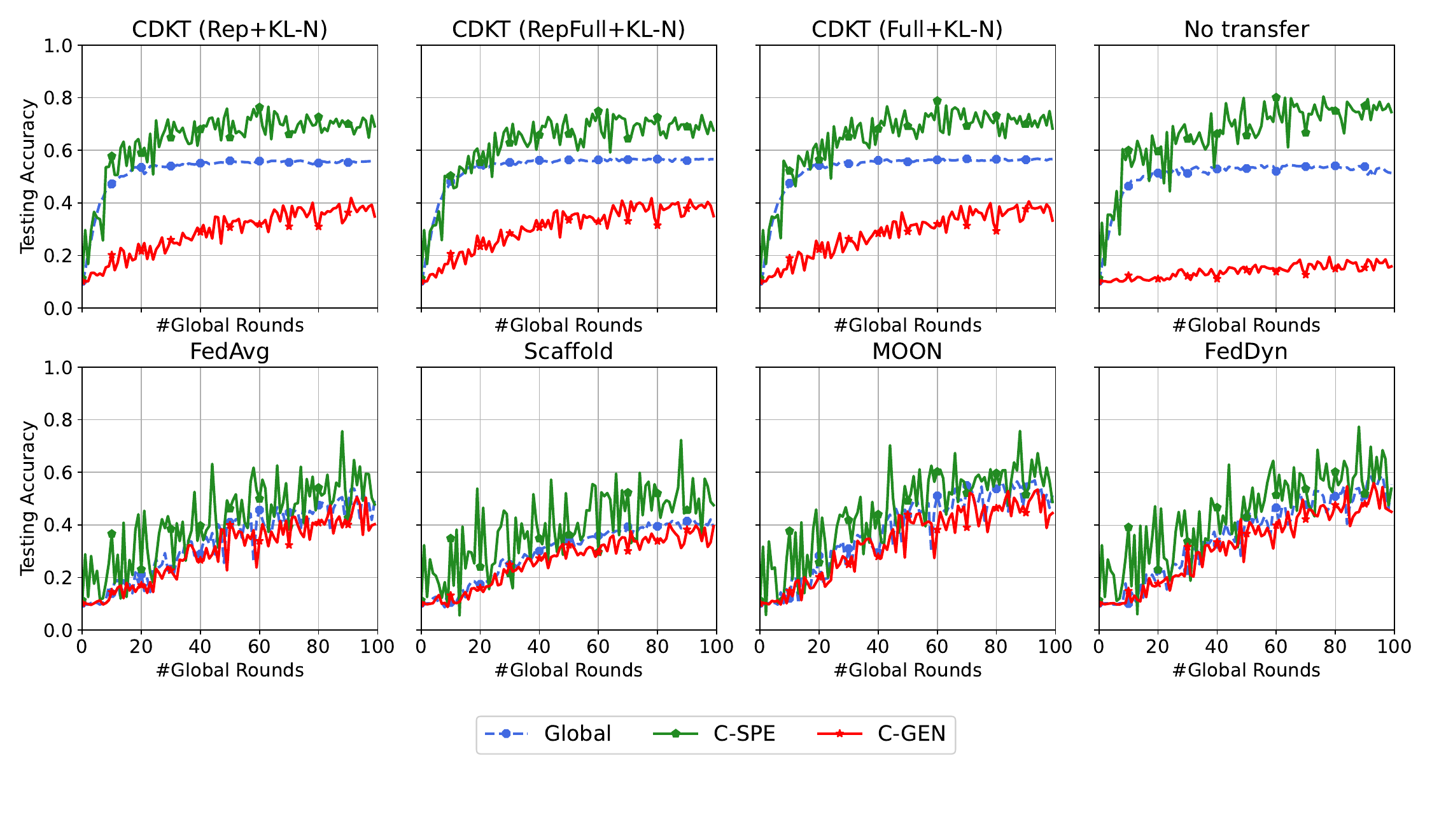}
	\caption{\textcolor{black}{Comparative analysis of \CDKT under various settings against other baselines using the CIFAR-10 dataset in Subset of Users scenario.}}
	\label{subset_users_cifar10}
\end{figure*}

\begin{figure*}[t]
	\centering
	\includegraphics[width=\linewidth]{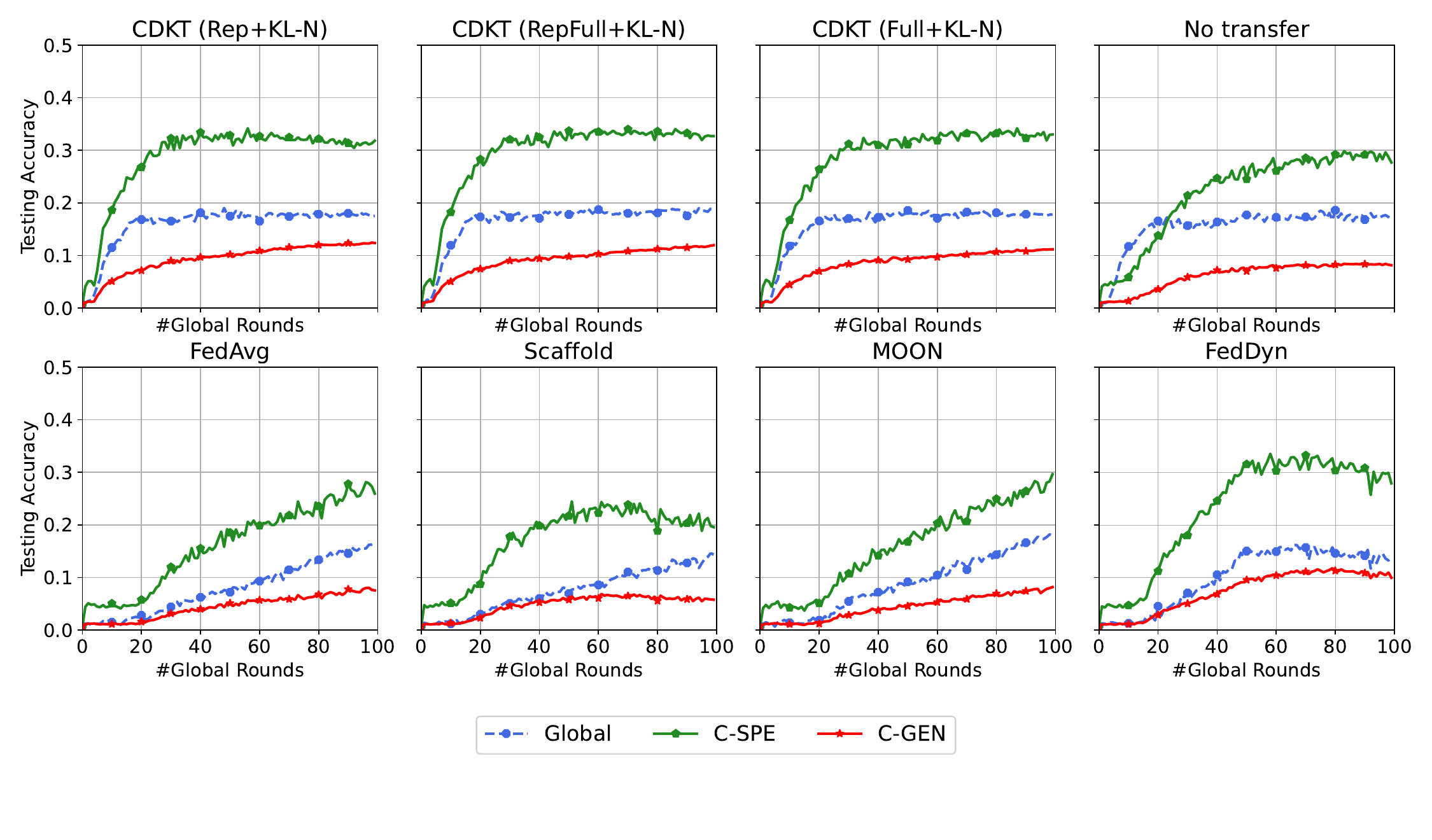}
	\caption{\textcolor{black}{Comparative analysis of \CDKT under various settings against other baselines using the CIFAR-100 dataset in Fixed Users scenario.}}
	\label{fixed_users_cifar100}
\end{figure*}

\begin{figure*}[t]
	\centering
	\includegraphics[width=\linewidth]{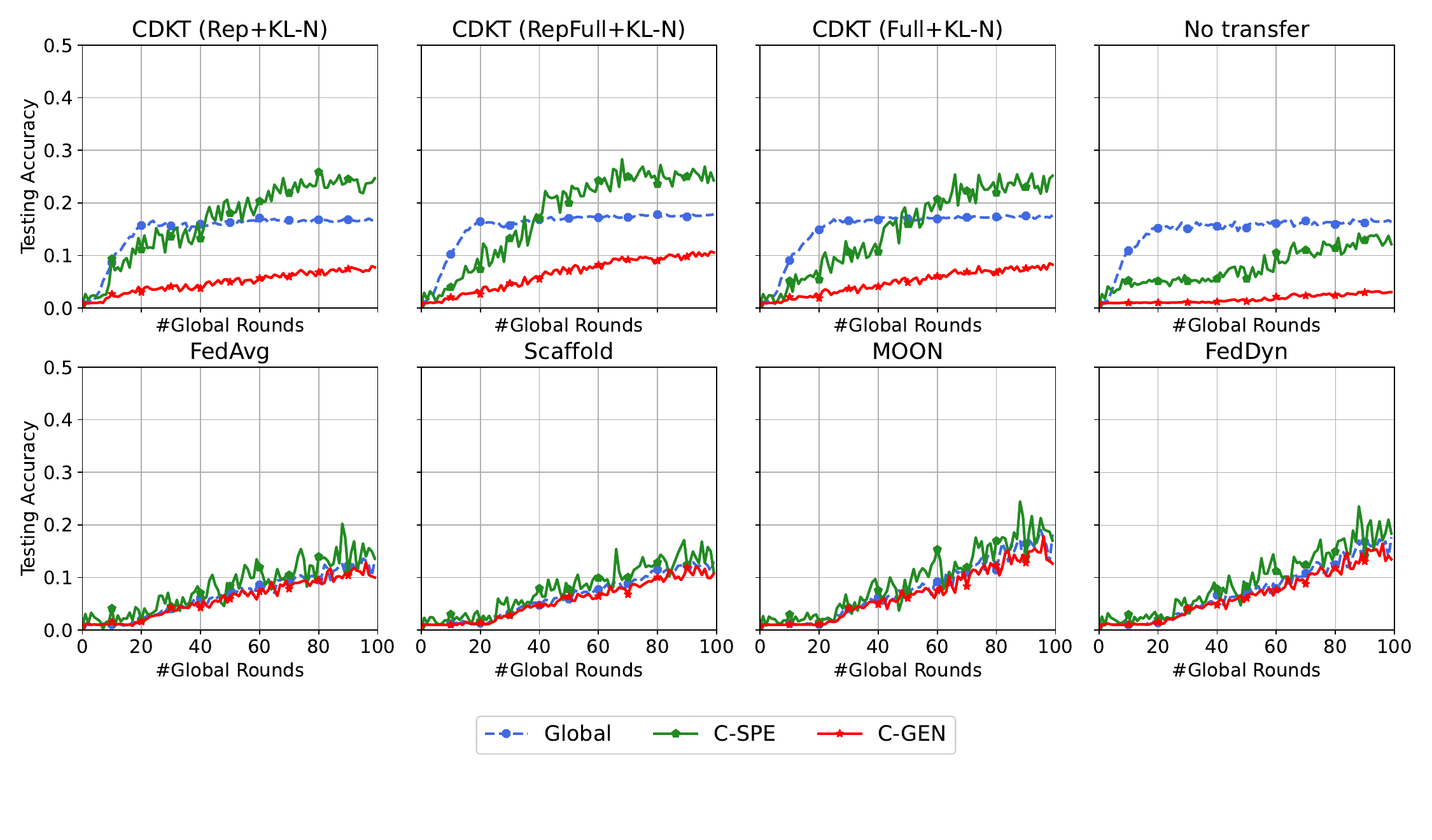}
	\caption{\textcolor{black}{Comparative analysis of \CDKT under various settings against other baselines using the CIFAR-100 dataset in Subset of Users scenario.}}
	\label{subset_users_cifar100}
\end{figure*}

\begin{figure*}
    \centering
    \begin{subfigure}{0.47\textwidth}
        \centering
        \includegraphics[width=1.1\linewidth]{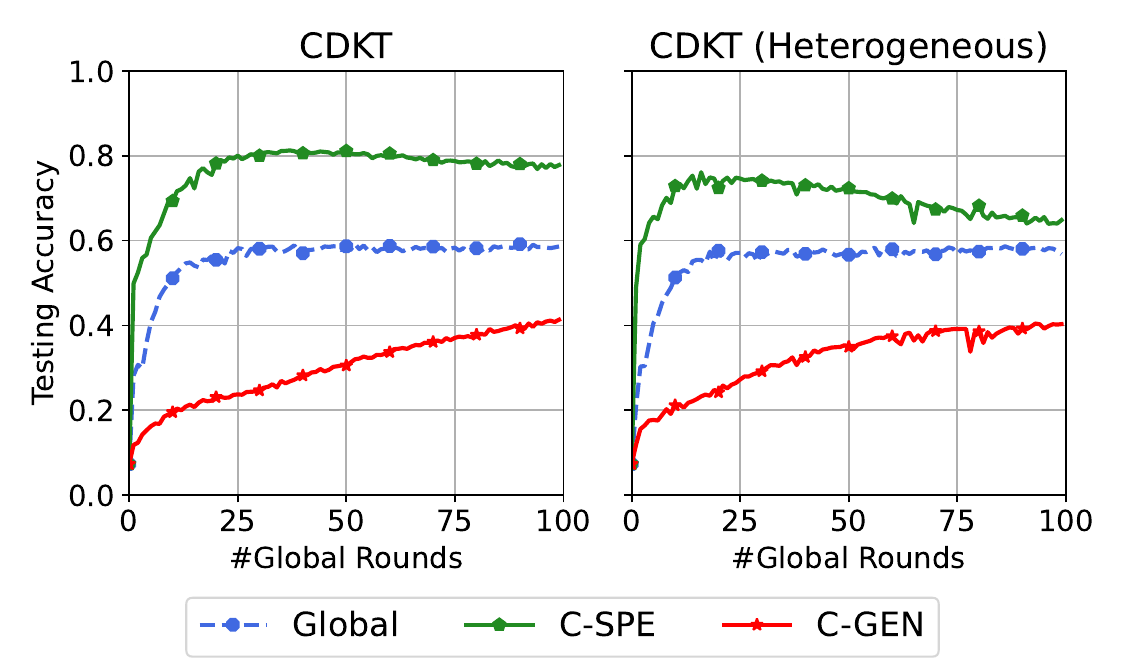}
        \caption{Fixed Users Scenario}
        \label{cifar10_metric_fixed}
    \end{subfigure}\hfill
    \begin{subfigure}{0.47\textwidth}
        \centering
        \includegraphics[width=1.1\linewidth]{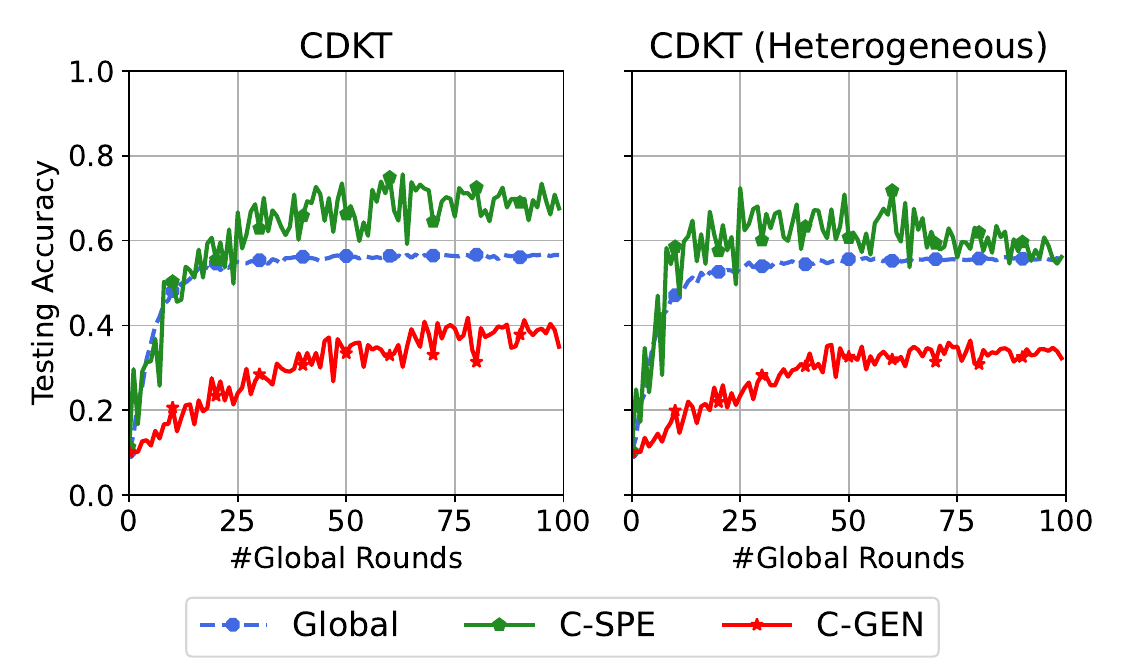}
        \caption{Subset of Users Scenario}
        \label{cifar10_metric_subset}
    \end{subfigure}
    \caption{Performance analysis of \CDKT with heterogeneous models on the CIFAR-10 dataset.}
    \label{cifar10_same_hetero}
\end{figure*}

\begin{figure*}
    \centering
    \begin{subfigure}{0.47\textwidth}
        \centering
        \includegraphics[width=1.1\linewidth]{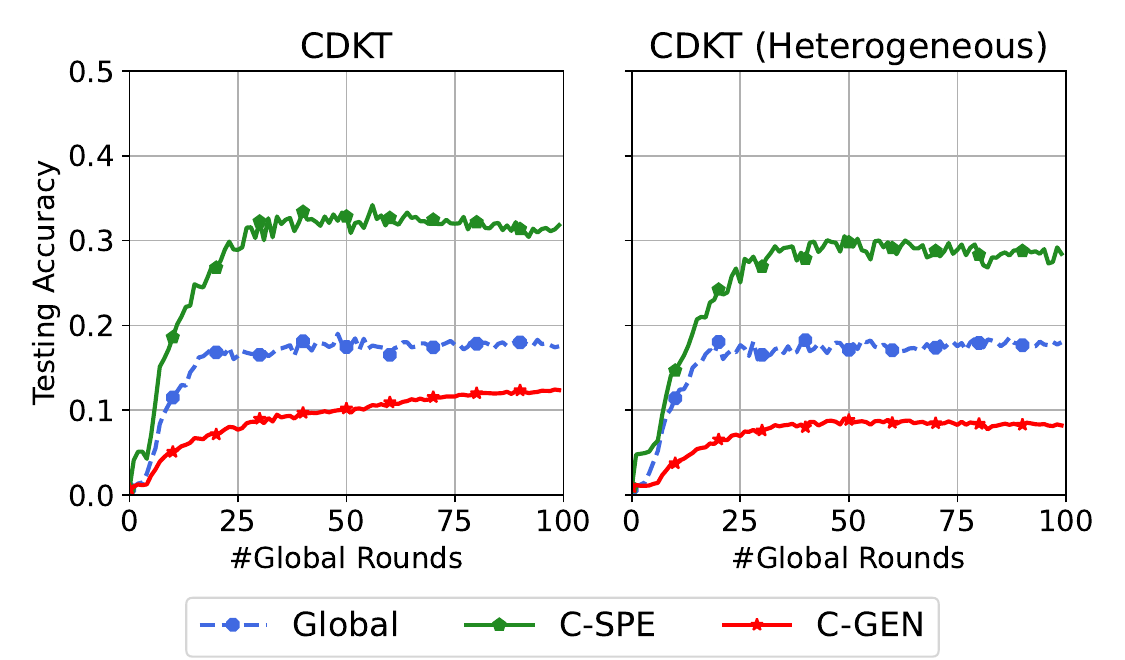}
        \caption{Fixed Users Scenario}
        \label{cifar100_metric_fixed}
    \end{subfigure}\hfill
    \begin{subfigure}{0.47\textwidth}
        \centering
        \includegraphics[width=1.1\linewidth]{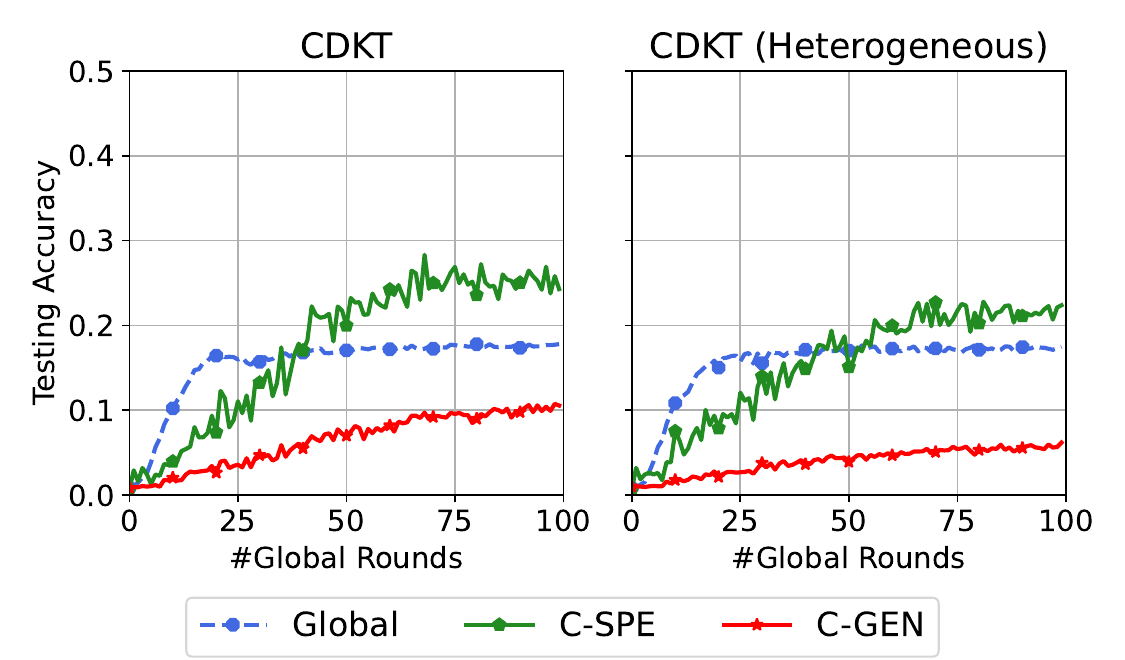}
        \caption{Subset of Users Scenario}
        \label{cifar100_metric_subset}
    \end{subfigure}
    \caption{Performance analysis of \CDKT with heterogeneous models on the CIFAR-100 dataset.}
    \label{cifar100_same_hetero}
\end{figure*}

\begin{figure*}[]
    \begin{subfigure}{\linewidth}
	\includegraphics[width=\linewidth]{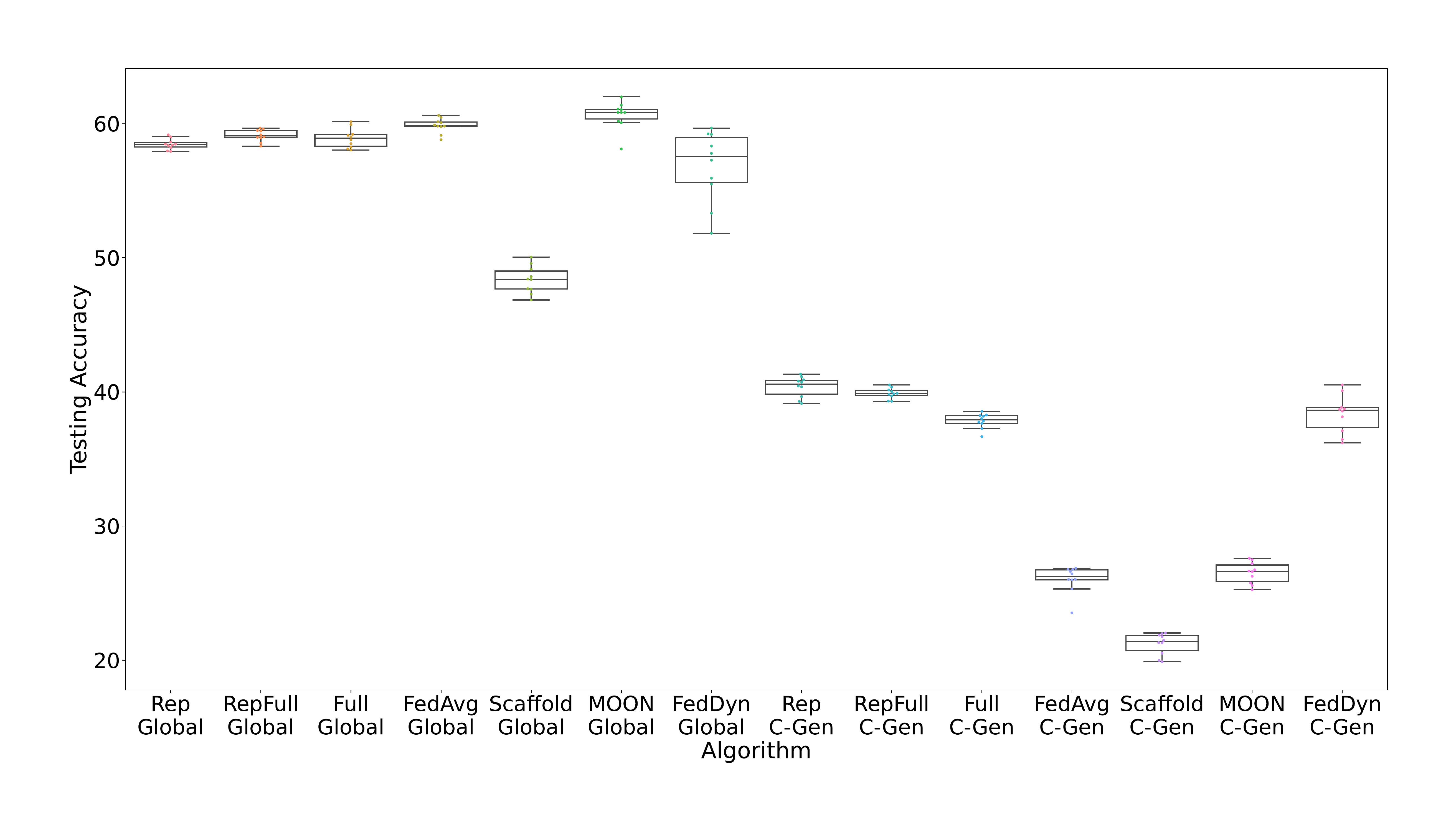}
	\caption{Fixed Users Scenario}
	\label{cifar10_fixed_box}
	\end{subfigure}
   \begin{subfigure}{\linewidth}
	\includegraphics[width=\linewidth]{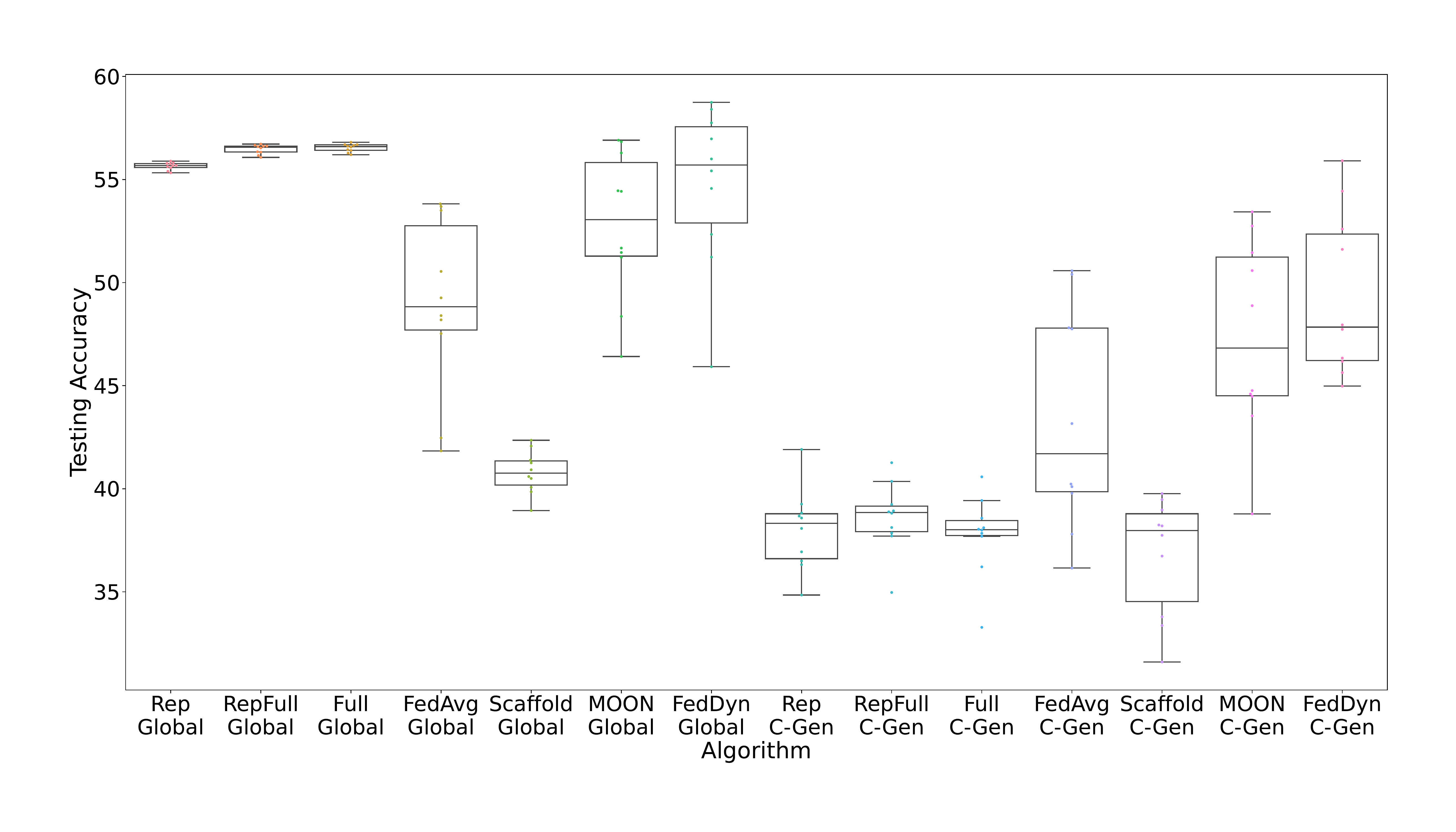}
	\caption{Subset of Users Scenario}
	\label{cifar10_subset_box}
	\end{subfigure}
	\caption{\textcolor{black}{Convergence analysis of \CDKT under various settings against other baselines using the CIFAR-10 dataset.}}
	\label{cifar10_box}
\end{figure*}

\begin{figure*}[]
    \begin{subfigure}{\linewidth}
	\includegraphics[width=\linewidth]{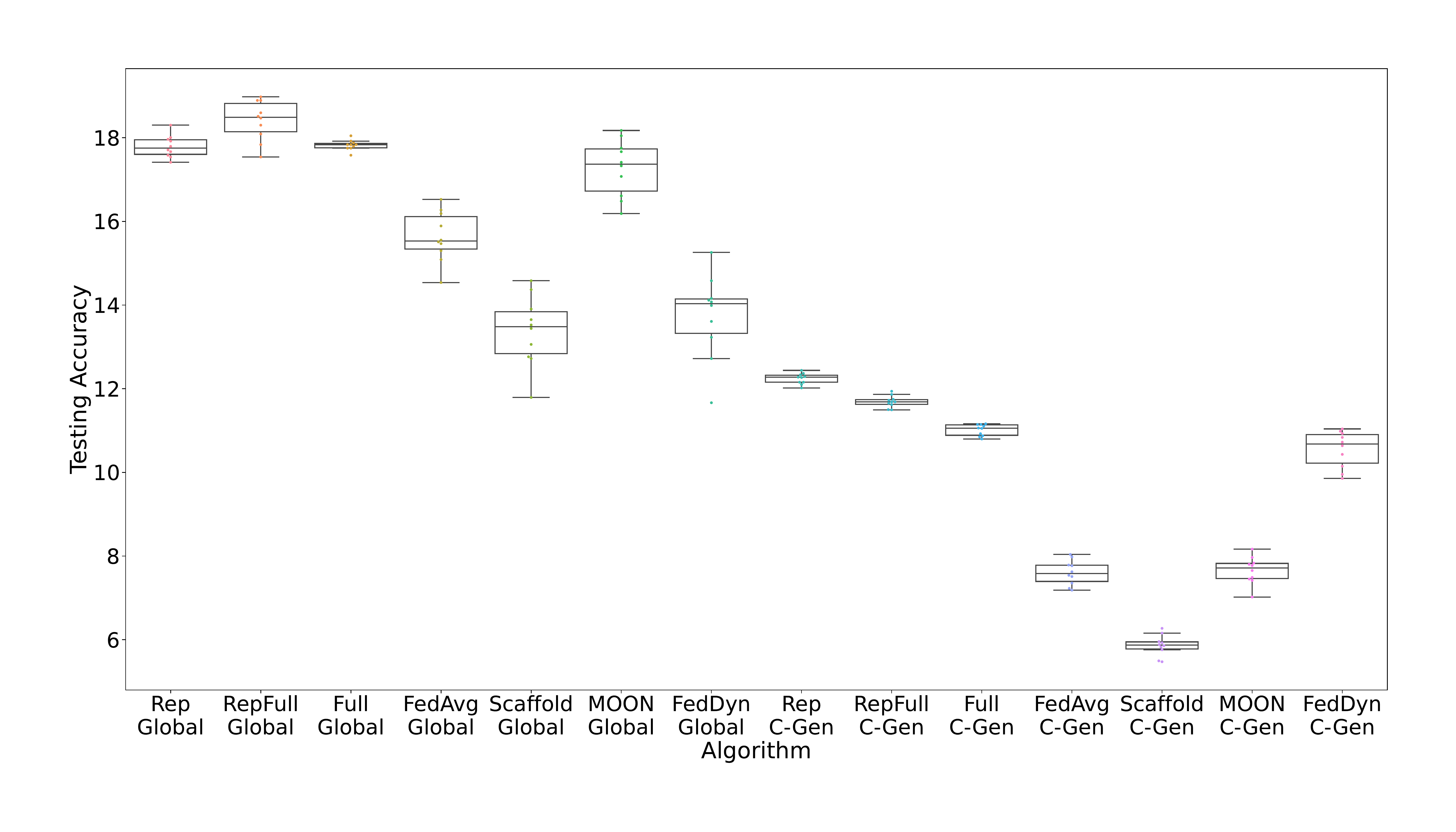}
	\caption{Fixed Users Scenario}
	\label{cifar100_fixed_box}
	\end{subfigure}
   \begin{subfigure}{\linewidth}
	\includegraphics[width=\linewidth]{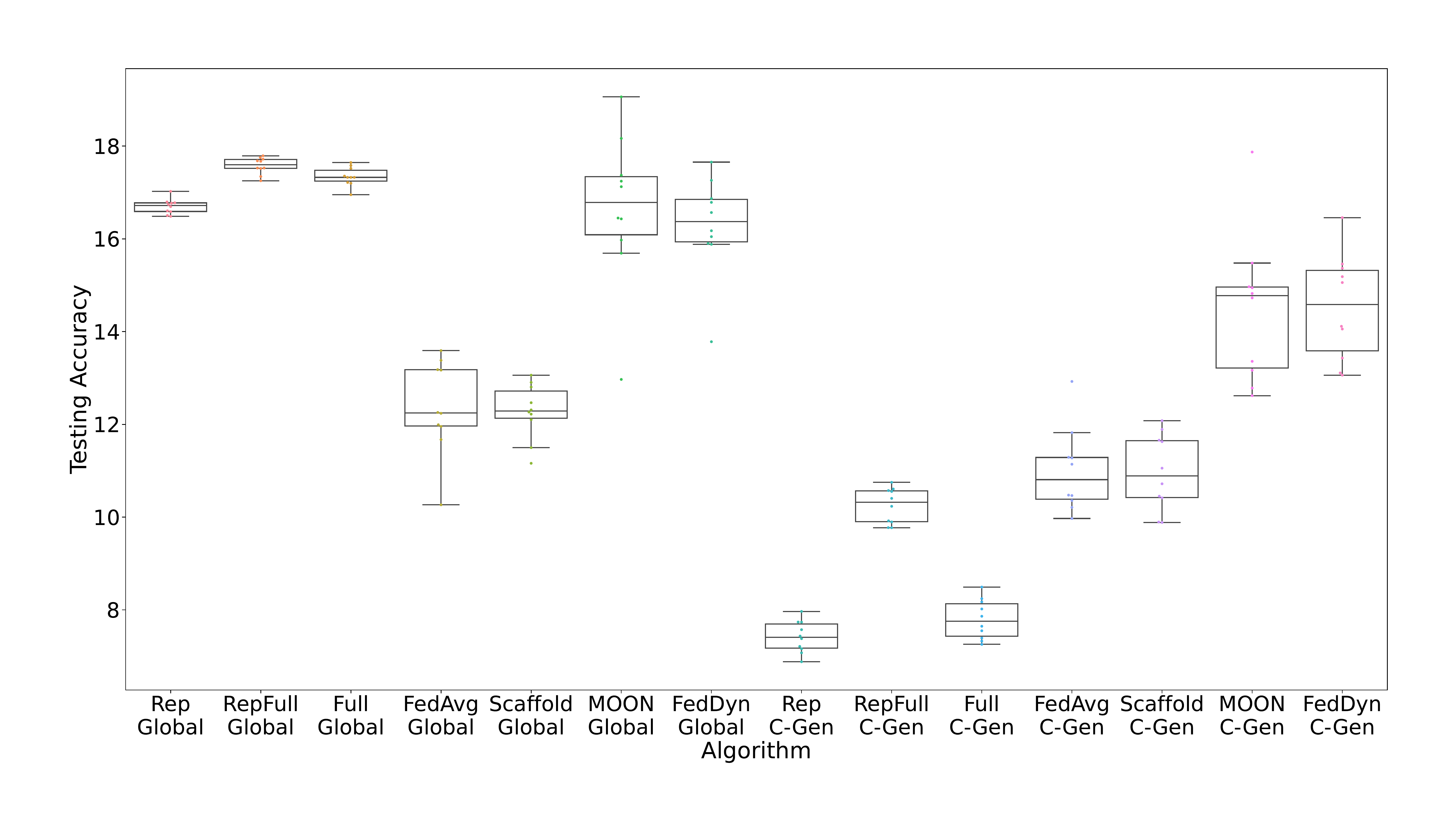}
	\caption{Subset of Users Scenario}
	\label{cifar100_subset_box}
	\end{subfigure}
	\caption{\textcolor{black}{Convergence analysis of \CDKT under various settings against other baselines using the CIFAR-100 dataset.}}
	\label{cifar100_box}
\end{figure*}

\begin{figure*}[ht]
  \subfloat[Fashion-MNIST]{
	\begin{minipage}[c][0.8\width]{
	   0.3\textwidth}
	   \centering
	   \includegraphics[width=1.1\textwidth]{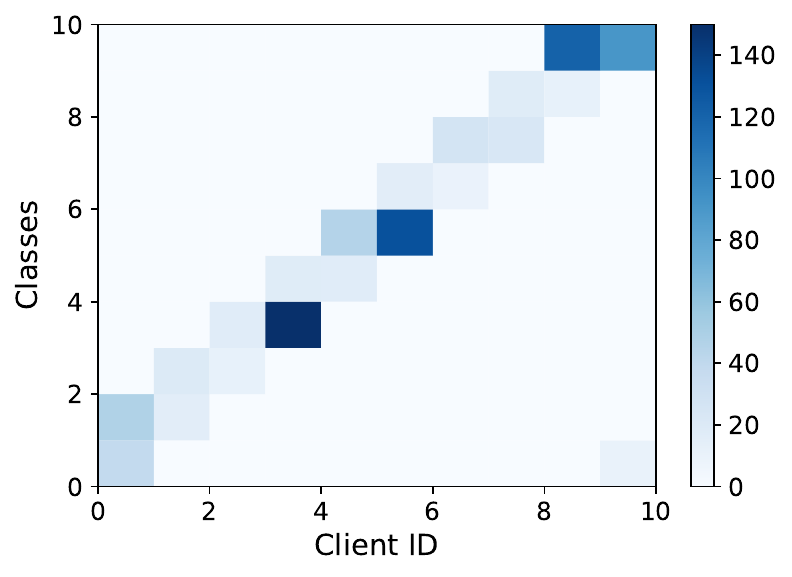}
	\end{minipage}}
 \hfill 	
  \subfloat[CIFAR-10]{
	\begin{minipage}[c][0.8\width]{
	   0.3\textwidth}
	   \centering
	   \includegraphics[width=1.1\textwidth]{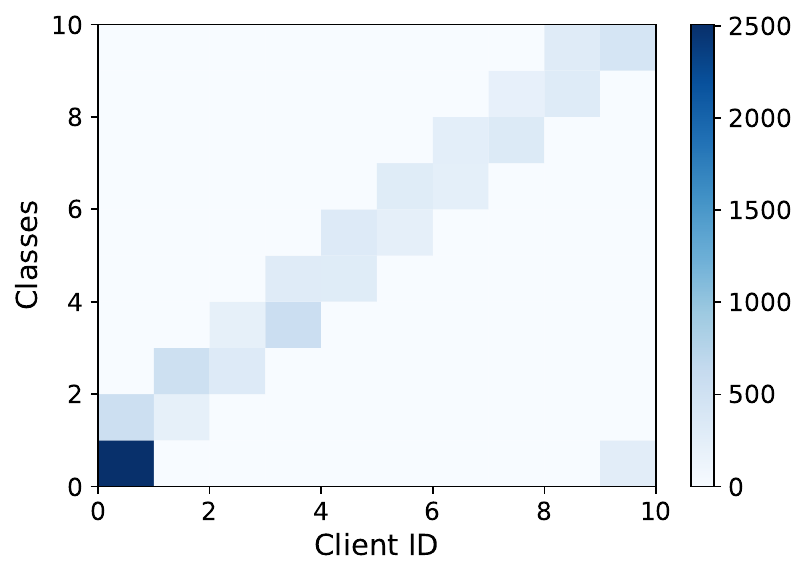}
	\end{minipage}}
 \hfill	
  \subfloat[CIFAR-100]{
	\begin{minipage}[c][0.8\width]{
	   0.3\textwidth}
	   \centering
	   \includegraphics[width=1.1\textwidth]{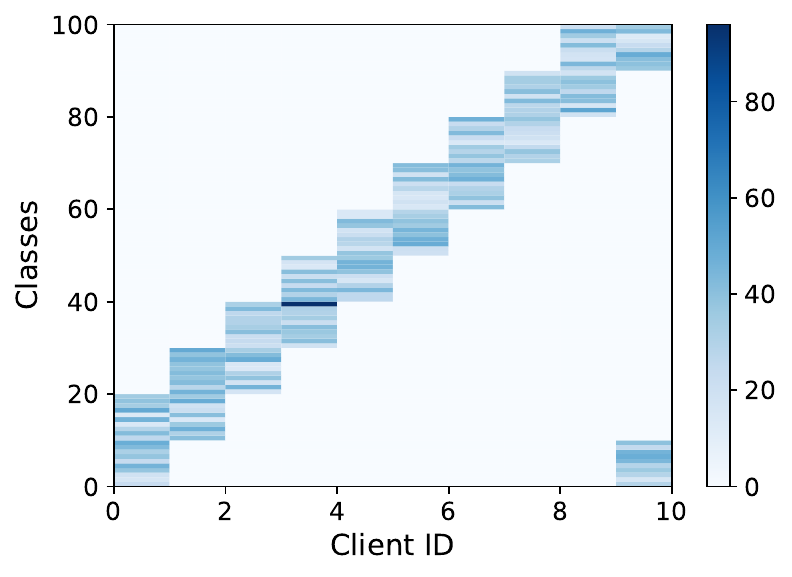}
	\end{minipage}}
\caption{\textcolor{black}{Visualization of the data distribution on three datasets for $10$ clients in the Fixed Users scenario. The color bar denotes the number of data samples. Each rectangle defines the number of samples in each class of each client.}}
\label{data_distribution_10clients}
\end{figure*}

\begin{figure*}[ht]
  \subfloat[Fashion-MNIST]{
	\begin{minipage}[c][0.8\width]{
	   0.3\textwidth}
	   \centering
	   \includegraphics[width=1.1\textwidth]{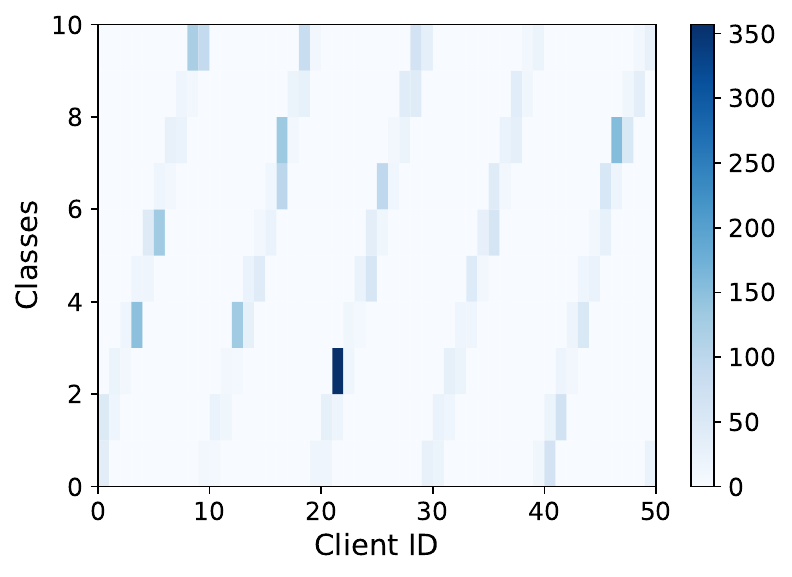}
	\end{minipage}}
 \hfill 	
  \subfloat[CIFAR-10]{
	\begin{minipage}[c][0.8\width]{
	   0.3\textwidth}
	   \centering
	   \includegraphics[width=1.1\textwidth]{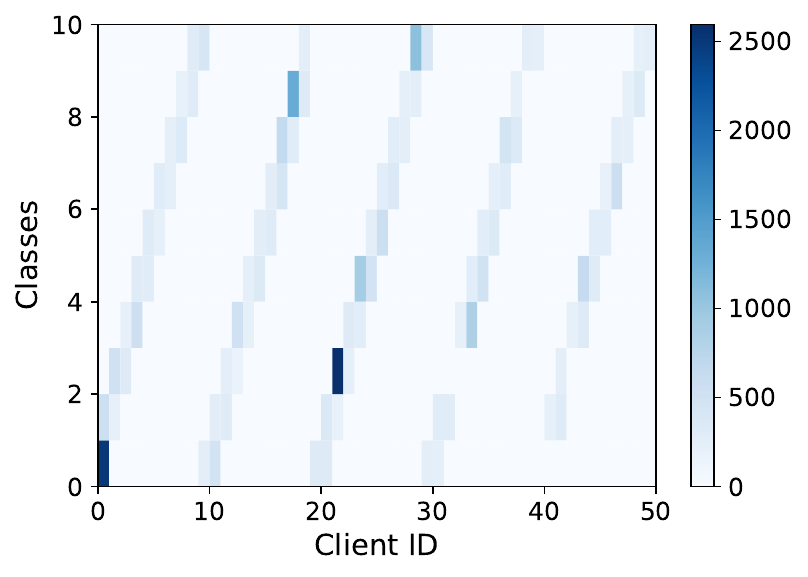}
	\end{minipage}}
 \hfill	
  \subfloat[CIFAR-100]{
	\begin{minipage}[c][0.8\width]{
	   0.3\textwidth}
	   \centering
	   \includegraphics[width=1.1\textwidth]{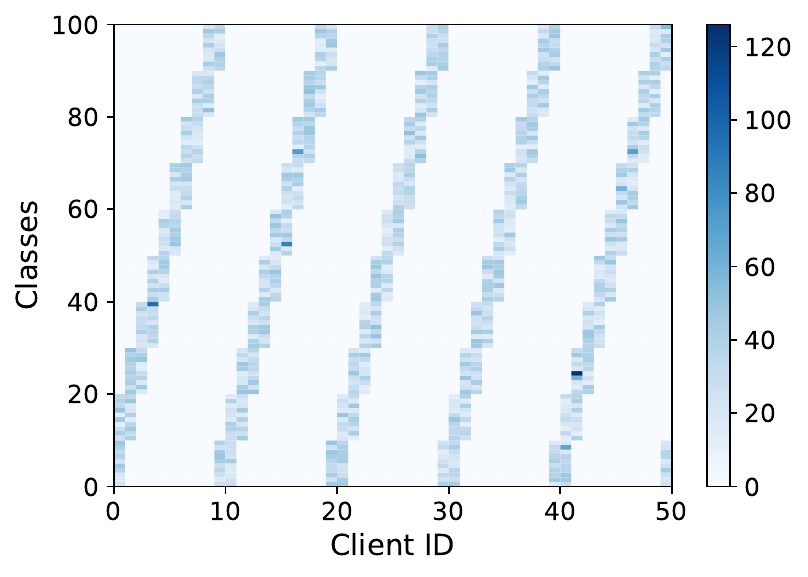}
	\end{minipage}}
\caption{\textcolor{black}{Visualization of the data distribution on three datasets for $50$ clients in the Subset of Users scenario. The color bar denotes the number of data samples. Each rectangle defines the number of samples in each class of each client.}}
\label{data_distribution}
\end{figure*}

\begin{figure*}[ht]
  \subfloat[Fashion-MNIST]{
	\begin{minipage}[c][0.8\width]{
	   0.3\textwidth}
	   \centering
	   \includegraphics[width=1.1\textwidth]{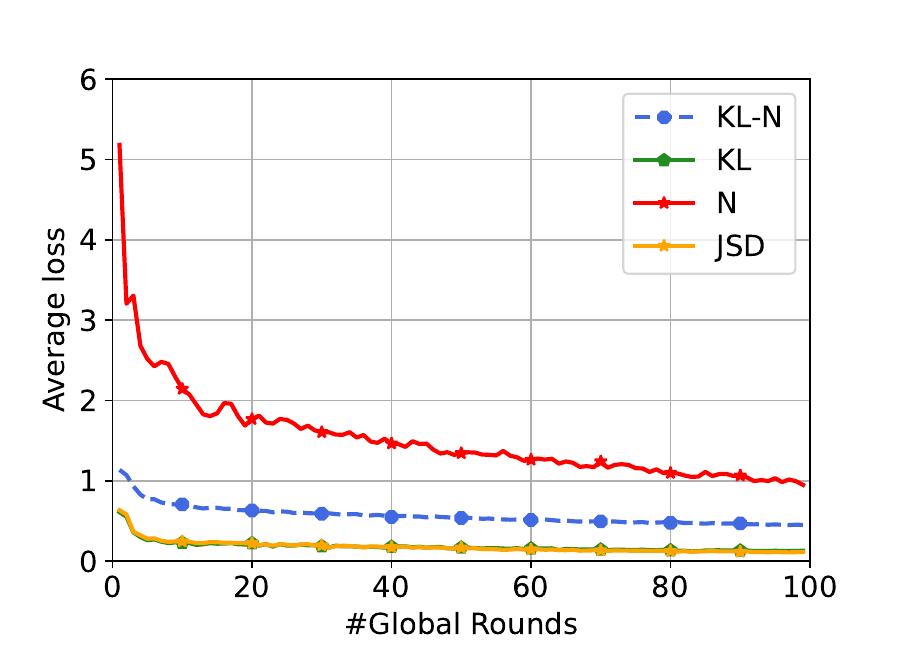}
	\end{minipage}}
 \hfill 	
  \subfloat[CIFAR-10]{
	\begin{minipage}[c][0.8\width]{
	   0.3\textwidth}
	   \centering
	   \includegraphics[width=1.1\textwidth]{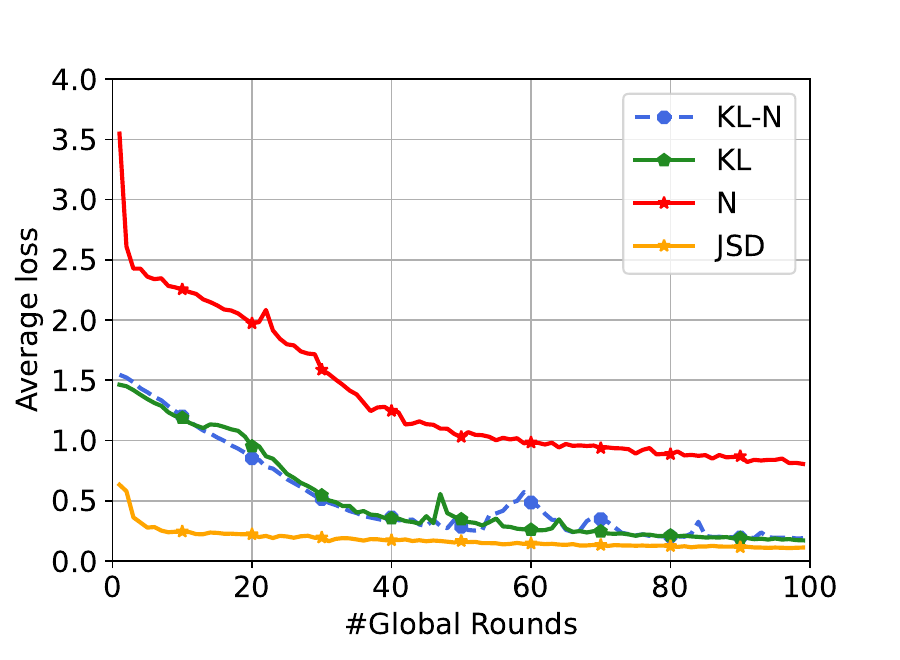}
	\end{minipage}}
 \hfill	
  \subfloat[CIFAR-100]{
	\begin{minipage}[c][0.8\width]{
	   0.3\textwidth}
	   \centering
	   \includegraphics[width=1.1\textwidth]{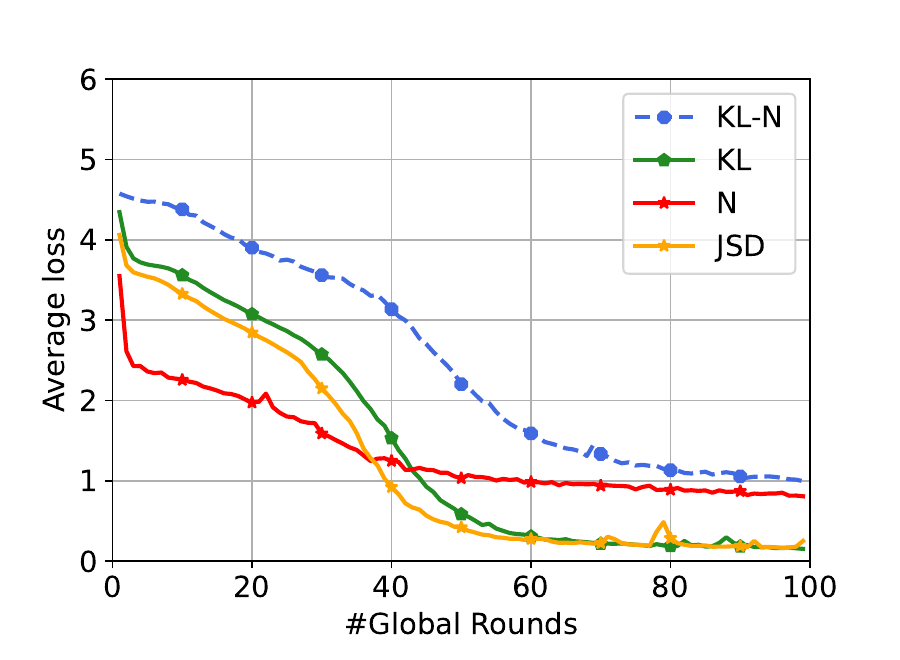}
	\end{minipage}}
\caption{\textcolor{black}{Average local loss versus global rounds on three datasets}}
\label{average_loss}
\end{figure*}

\begin{figure*}[ht]
  \subfloat[Fashion-MNIST]{
	\begin{minipage}[c][0.8\width]{
	   0.3\textwidth}
	   \centering
	   \includegraphics[width=1.1\textwidth]{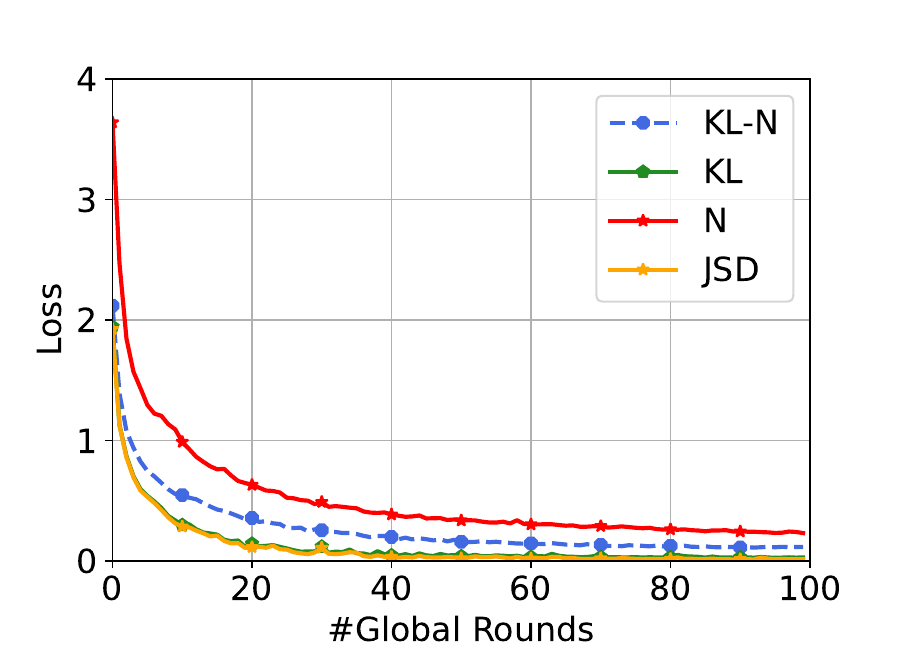}
	\end{minipage}}
 \hfill 	
  \subfloat[CIFAR-10]{
	\begin{minipage}[c][0.8\width]{
	   0.3\textwidth}
	   \centering
	   \includegraphics[width=1.1\textwidth]{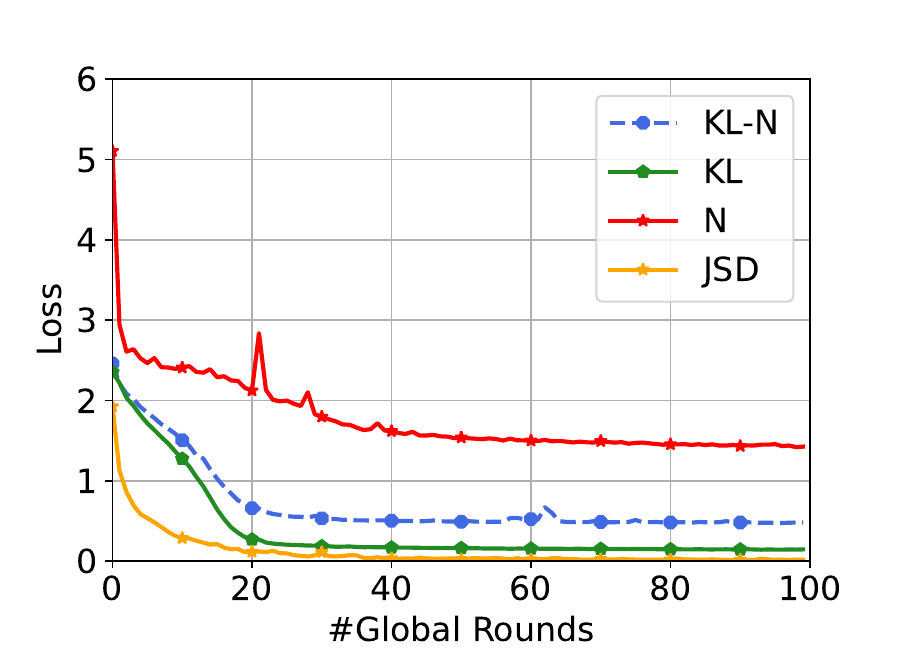}
	\end{minipage}}
 \hfill	
  \subfloat[CIFAR-100]{
	\begin{minipage}[c][0.8\width]{
	   0.3\textwidth}
	   \centering
	   \includegraphics[width=1.1\textwidth]{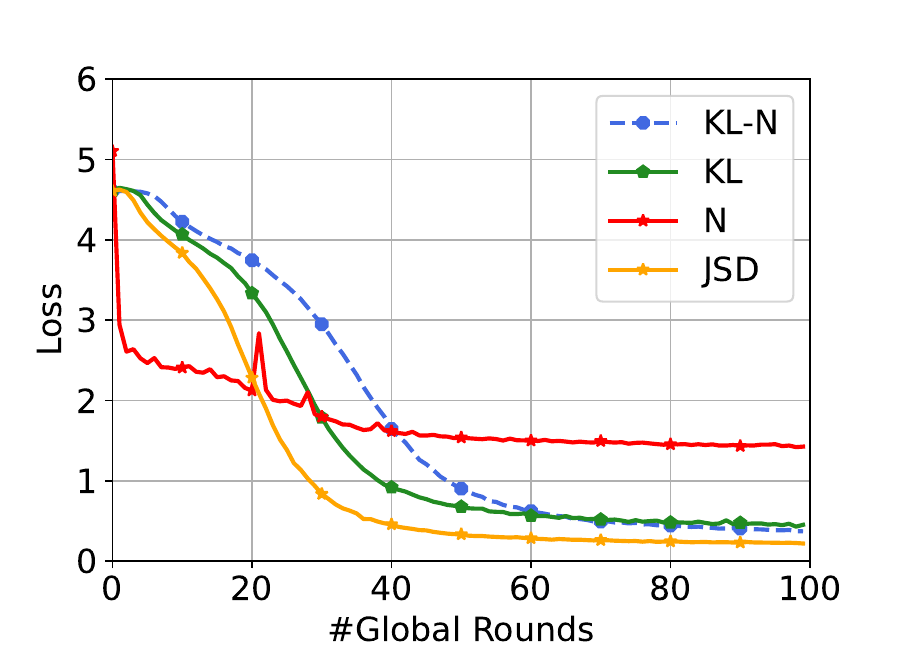}
	\end{minipage}}
\caption{\textcolor{black}{Global loss versus global rounds on three datasets}}
\label{global_loss}
\end{figure*}

\begin{figure*}[]
 \centering
    \begin{subfigure}{\linewidth}
	\includegraphics[width=\linewidth]{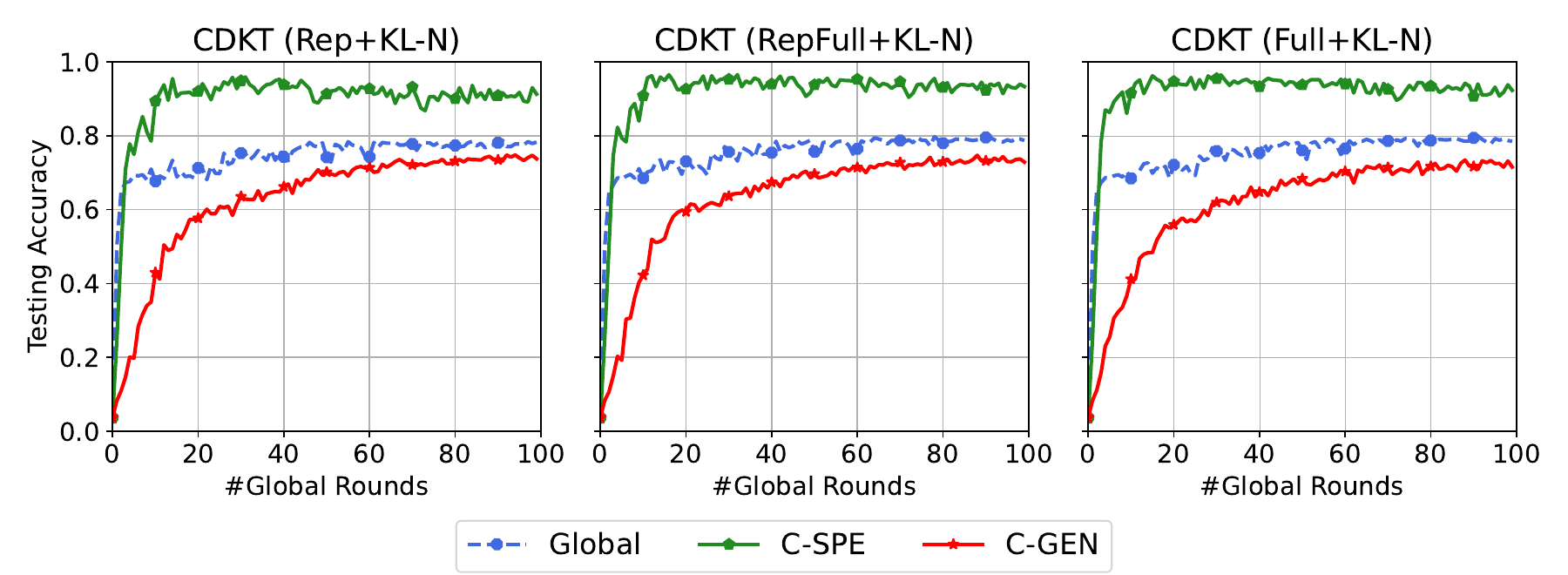}
	\caption{50 clients (sample fraction=$0.5$)}
	\label{fmnist_fixed_box}
    \end{subfigure}
    \begin{subfigure}{\linewidth}
	\includegraphics[width=\linewidth]{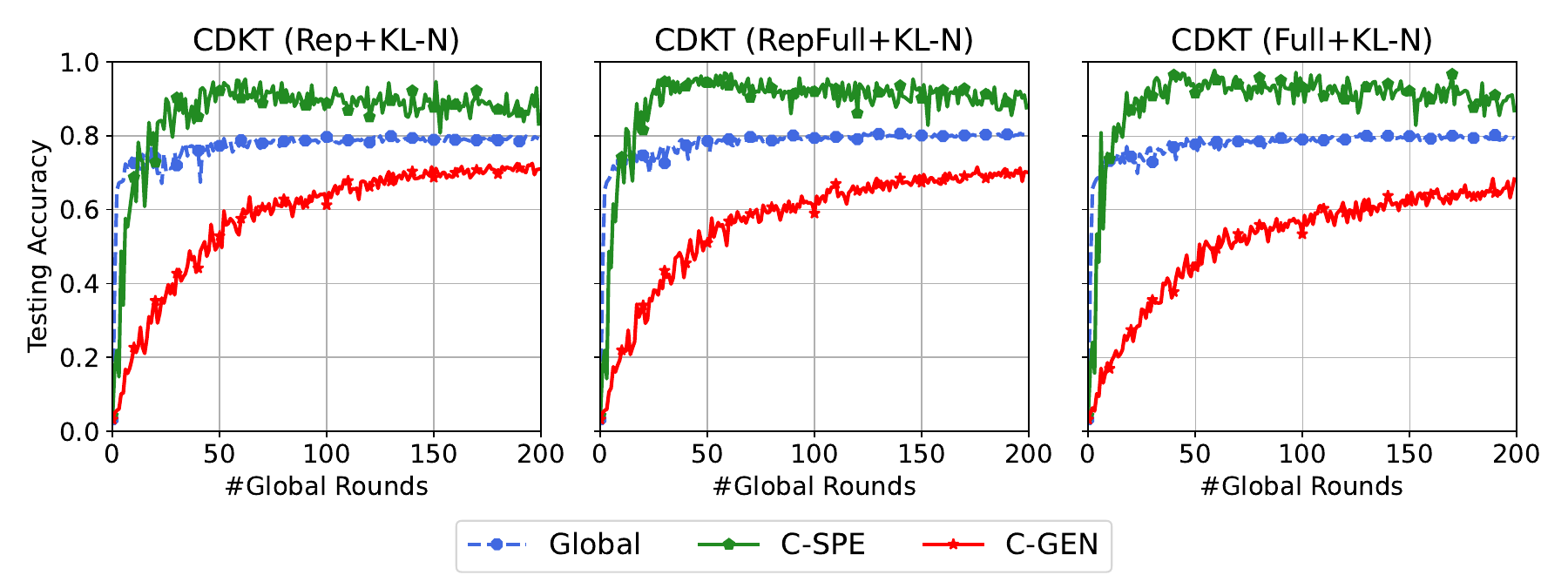}
	\caption{100 clients (sample fraction=$0.2$)}
	\label{fmnist_subset_box}
	\end{subfigure}
	\caption{\textcolor{black}{Convergence of \CDKT with different settings of clients Fashion-MNIST dataset}}
	\label{fmnist_scalability}
\end{figure*}

\begin{figure*}[]
 \centering
    \begin{subfigure}{\linewidth}
	\includegraphics[width=\linewidth]{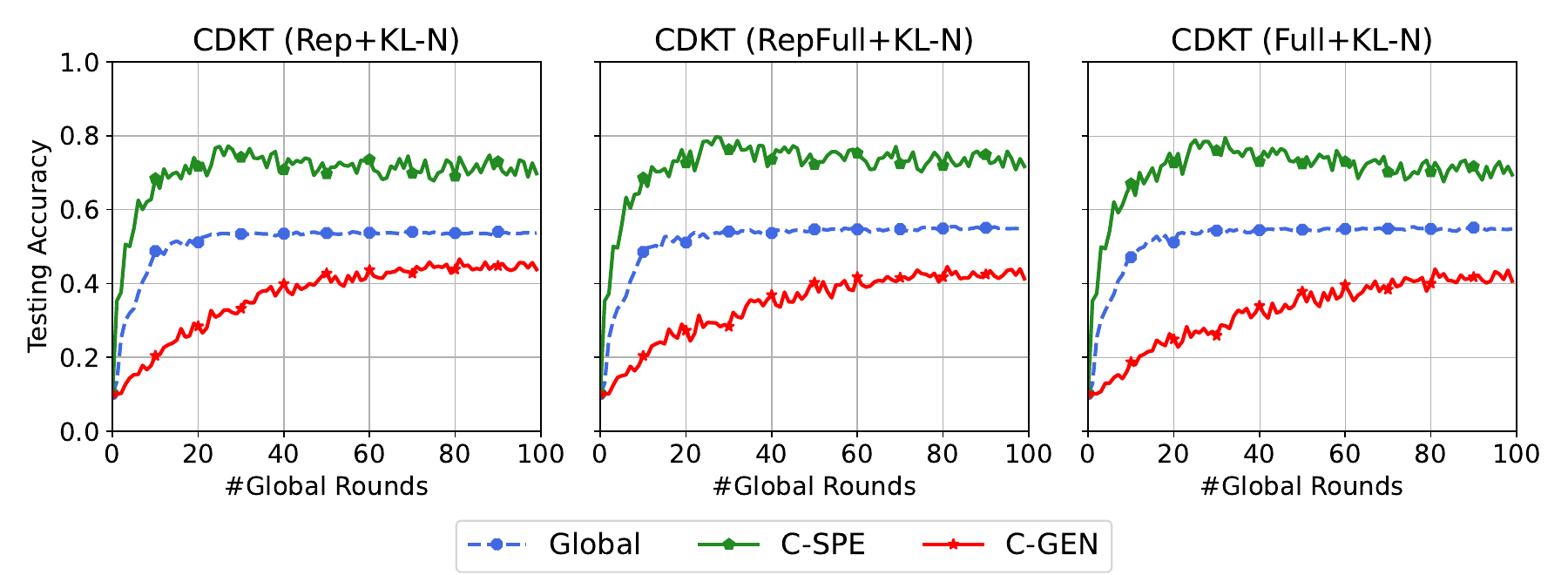}
	\caption{50 clients (sample fraction=$0.5$)}
	\label{fmnist_fixed_box}
    \end{subfigure}
    \begin{subfigure}{\linewidth}
	\includegraphics[width=\linewidth]{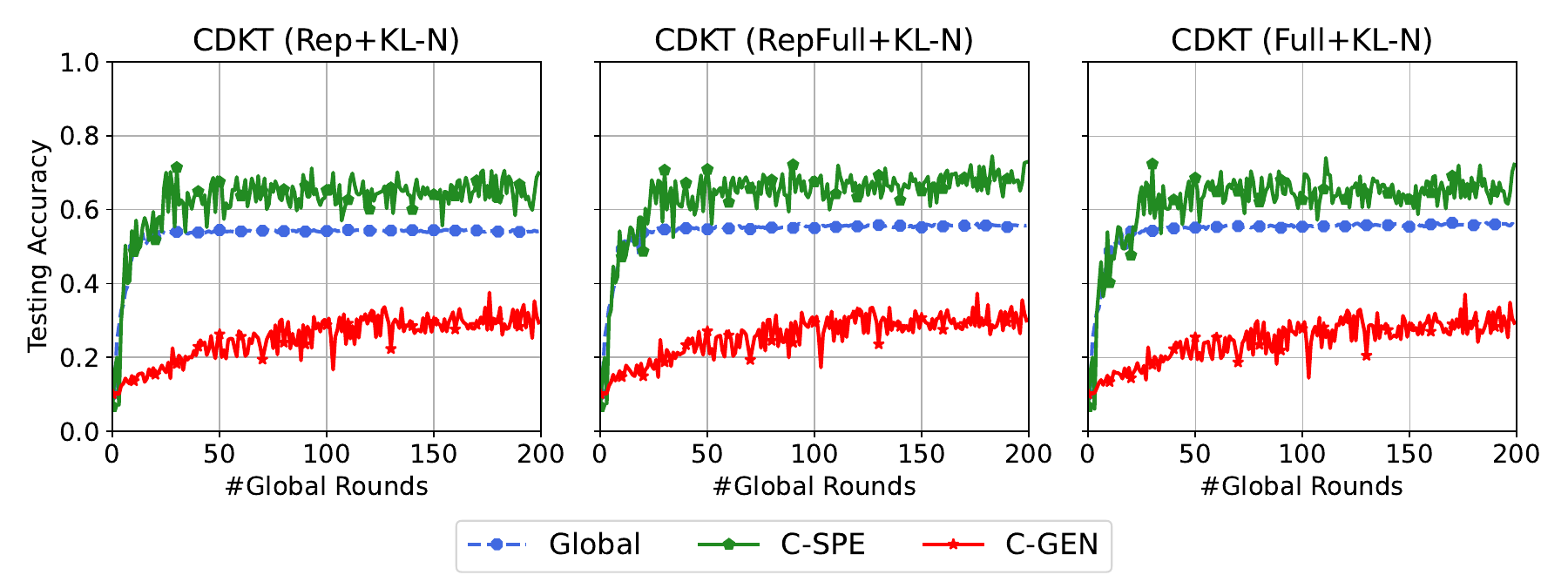}
	\caption{100 clients (sample fraction=$0.2$)}
	\label{fmnist_subset_box}
	\end{subfigure}
	\caption{\textcolor{black}{Convergence of \CDKT with different settings of clients with CIFAR-10 dataset}}
	\label{cifar10_scalability}
\end{figure*}

\begin{figure*}[]
 \centering
    \begin{subfigure}{\linewidth}
	\includegraphics[width=\linewidth]{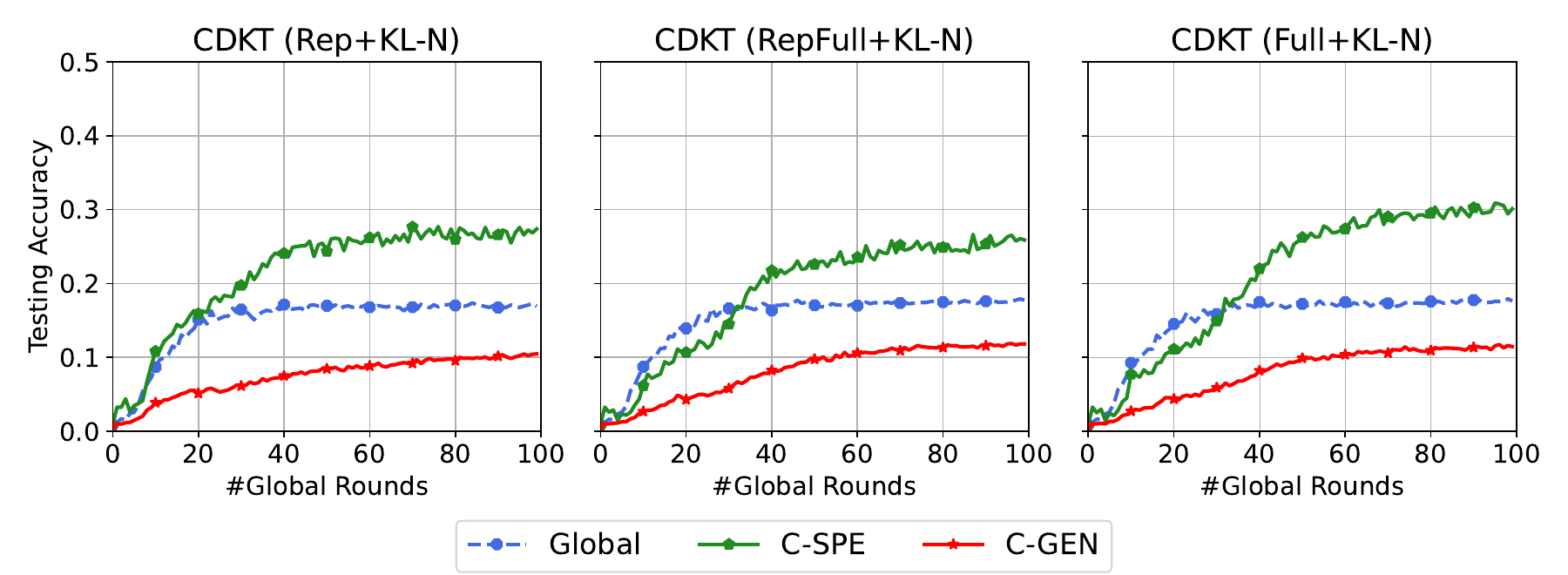}
	\caption{50 clients (sample fraction=$0.5$)}
	\label{fmnist_fixed_box}
    \end{subfigure}
    \begin{subfigure}{\linewidth}
	\includegraphics[width=\linewidth]{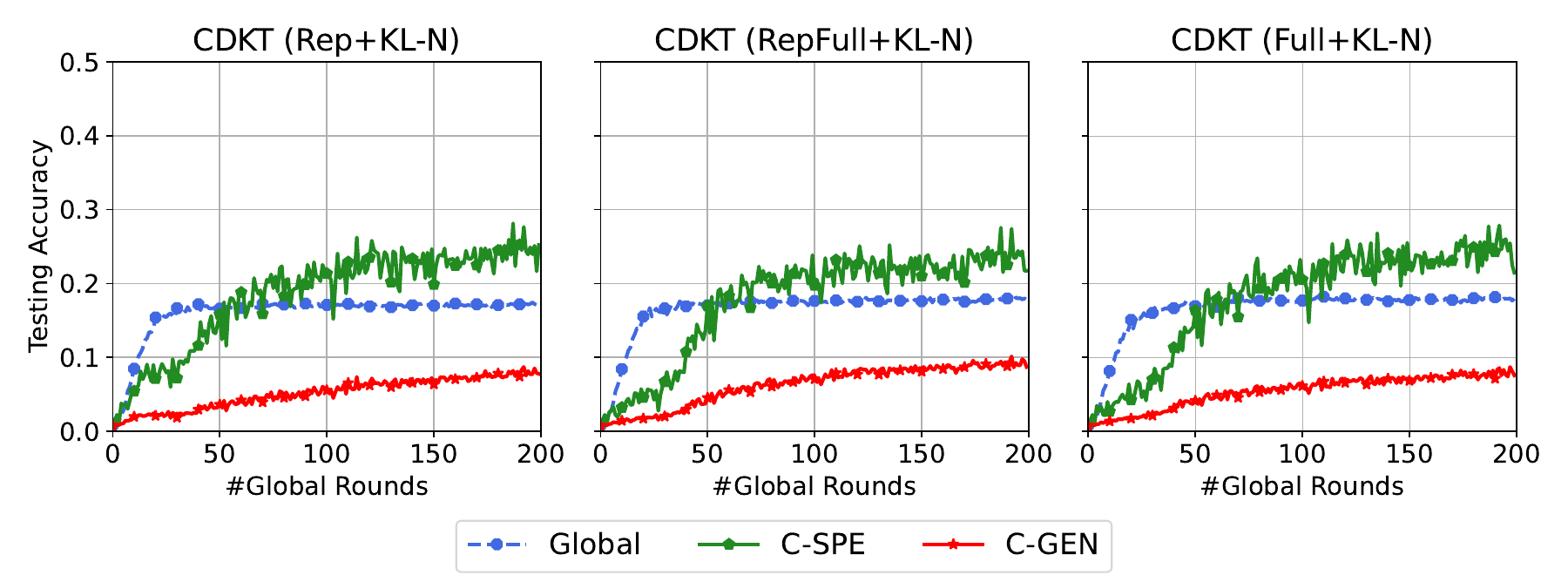}
	\caption{100 clients (sample fraction=$0.2$)}
	\label{fmnist_subset_box}
	\end{subfigure}
	\caption{\textcolor{black}{Convergence of \CDKT with different settings of clients with CIFAR-100 dataset}}
	\label{cifar100_scalability}
\end{figure*}

 \textcolor{black}{Similar to the learning behaviors with the Fashion-MNIST dataset in the main manuscript, CIFAR-10 and CIFAR-100 exhibit faster convergence of the \textbf{C-Gen} performance, comparable \textbf{Global} and \textbf{C-Spec} performance and its stability with different settings as shown in Fig.~\ref{fixed_users_cifar10} and Fig.~\ref{fixed_users_cifar100}, respectively, in scenario (1). Regarding scenario (2), as shown in Fig.~\ref{subset_users_cifar10} and Fig.~\ref{subset_users_cifar100}, FedDyn and MOON depict the highest \textbf{C-Gen} performance in two datasets respectively, meanwhile \CDKT outperforms all baselines in terms of \textbf{Global} and \textbf{C-Spec} performance. Also, other methods extensively fluctuate since the learned models differ highly between clients. Moreover, as shown in Fig.~\ref{cifar10_same_hetero} and Fig.~\ref{cifar100_same_hetero}, using smaller learning models at clients with two CNN layers while three CNN layers in the server model show a slight degradation due to lower learning capacities in both datasets. Additionally, we plot the box plot for different settings of \CDKT to compare with other methods in terms of stability in both scenarios, as shown in Fig.~\ref{cifar10_box} and Fig.~\ref{cifar100_box}. Similar to the Fashion-MNIST dataset, our algorithm achieves a stable convergence in the last 20 rounds, while other baselines show significant fluctuation. }
\textcolor{black}{\subsection{Model Architecture}
In Table~\ref{model_architecture_fashion} and Table~\ref{model_architecture_cifar}, we provide a summary of model architecture that we used in our experiments. For Fashion-MNIST, we use the CNN model with two convolution layers, where as three convolution layers are used in CIFAR-10 and CIFAR-100 datasets, followed by max pooling and fully connected layers.} 
\textcolor{black}{\subsection{Data Partition}
We provide the data partition of three datasets in Fig.~\ref{data_distribution_10clients} and Fig.~\ref{data_distribution}. We generate the non-i.i.d properties by assigning data from randomly $2$ classes out of the total classes to each client for Fashion-MNIST and CIFAR-10 datasets and 20 classes in 100 classes for CIFAR-100 datasets.}
\textcolor{black}{\subsection{Detailed of C-Gen and C-Spec performance}
We provide the detailed results of average accuracy and F1-score of C-Gen and C-Spec performance for scenario (1) in Table~\ref{appendix_fixed} and Table~\ref{appendix_fixed_f1} and scenario (2) in Table~\ref{appendix_subset} and Table~\ref{appendix_subset_F1}.}
\textcolor{black}{\subsection{Loss vs. Global Rounds Analysis}
\textcolor{black}{We analyze the impact of different loss functions on the training process and model performance regarding average local loss and global loss on three datasets, as shown in Fig.~\ref{average_loss} and Fig.~\ref{global_loss}. We observe that JSD and KL converge quickly in terms of average local loss, whereas KL-N shows the best convergence in terms of global loss. In both two cases, N performs the slowest decline. In our experiments, although JSD illustrates the solid convergence in terms of loss value, KL-N exhibits the best accuracy in most scenarios.}}
\textcolor{black}{\subsection{Scalability}
To assess the impact of increasing the number of local users, we conducted experiments under two settings: (1) partitioning the dataset into 50 clients and randomly sampling 25 clients for each training round, and (2) partitioning the dataset into 100 clients with a random sample of 20 clients per training round. The median testing accuracy for these scenarios is presented in Table~\ref{scalability_fmnist}, Table~\ref{scalability_cifar10} and Table~\ref{scalability_cifar100}, while the convergence in the number of communication rounds is depicted in Fig.~\ref{fmnist_scalability}, Fig.~\ref{cifar10_scalability} and Fig.~\ref{cifar100_scalability}. The results highlight distinct patterns in \textbf{Global} and \textbf{C-Per} performance across all datasets. In terms of \textbf{Global} performance, an increase in the sample fraction from $0.2$ to $0.5$ prompts the global model to generalize across a larger set of clients, leading to a slight decrease in performance. On the other hand, an increase in the number of clients results in comparable \textbf{Global} performance after a greater number of communication rounds. Turning to \textbf{C-Per} performance, elevating the sample fraction from $0.2$ to $0.5$ tends to enhance \textbf{C-Per} outcomes as the subset of clients undergoes fewer replacements. Increasing the sample fraction provides clients with more opportunities to participate in the training process, thereby leading to improved \textbf{C-Per} performance. However, increasing the number of clients with a consistent fraction leads to a decline in \textbf{C-Per} performance, attributed to the more frequent replacement of the subset of clients. These findings affirm the scalability of our approach and its ability to handle diverse numbers of local devices while preserving performance.
}

\begin{table*}[]
\begin{center}
\textcolor{black}{
\caption{\textcolor{black}{Detailed information of the CNN architecture used in our experiment for Fashion-MNIST dataset.}}
\label{model_architecture_fashion}
\begin{tabular}{|l|l|}
\hline
\textbf{Layer} & \textbf{Parameter  \& Shape \& Hyper-parameters}                   \\ \hline
conv1          & Conv2d(1, 32, kernel\_size=(5, 5), stride=(1, 1), padding=(2, 2))  \\ \hline
pool1          & MaxPool2d(kernel\_size=2, stride=2, padding=0, dilation=1)         \\ \hline
dropout1       & Dropout(p=0.4)                                                     \\ \hline
conv2          & Conv2d(32, 32, kernel\_size=(5, 5), stride=(1, 1), padding=(2, 2)) \\ \hline
pool2          & MaxPool2d(kernel\_size=2, stride=2, padding=0, dilation=1)         \\ \hline
dropout2       & Dropout(p=0.4)                                                     \\ \hline
fc1            & Linear(in\_features=32*7*7, out\_features=512)                     \\ \hline
fc2            & Linear(in\_features=512, out\_features=10)                         \\ \hline
\end{tabular}
}
\end{center}
\end{table*}

\begin{table*}[]
\begin{center}
\textcolor{black}{
\caption{\textcolor{black}{Detailed information of the CNN architecture used in our experiment for CIFAR-10 and CIFAR-100 datasets.}}
\label{model_architecture_cifar}
\begin{tabular}{|l|l|}
\hline
\textbf{Layer} & \textbf{Parameter  \& Shape \& Hyper-parameters}                   \\ \hline
conv1          & Conv2d(3, 16, kernel\_size=(3,3), stride=(1,1), padding=(1, 1))    \\ \hline
pool1          & MaxPool2d(kernel\_size=2, stride=2, padding=0, dilation=1)         \\ \hline
conv2          & Conv2d(16, 32, kernel\_size=(3, 3), stride=(1, 1), padding=(1, 1)) \\ \hline
pool2          & MaxPool2d(kernel\_size=2, stride=2, padding=0, dilation=1)         \\ \hline
conv3          & Conv2d(32, 64, kernel\_size=(3, 3), stride=(1, 1), padding=(1, 1)) \\ \hline
pool3          & MaxPool2d(kernel\_size=2, stride=2, padding=0, dilation=1)         \\ \hline
fc1            & Linear(in\_features=64*4*4, out\_features=512)          \\ \hline
fc2            & Linear(in\_features=512, out\_features=64)              \\ \hline
dropout        & Dropout(p=0.5)                                                     \\ \hline
fc3            & Linear(in\_features=64, out\_features=num\_classes, bias=True)     \\ \hline
\end{tabular}
}
\end{center}
\end{table*}


\begin{table}[]
\caption{\textcolor{black}{The median of test accuracy across three datasets over a range of $90$ to $100$ rounds in Fixed Users scenario.}}
\label{appendix_fixed}
\begin{center}
\resizebox{14cm}{!}{%
\textcolor{black}{
\begin{tabular}{|c|ccc|ccc|ccc|}
\hline
\multicolumn{1}{|l|}{\multirow{2}{*}{}}                           & \multicolumn{3}{c|}{\textbf{Fashion-MNIST}}                                                 & \multicolumn{3}{c|}{\textbf{CIFAR-10}}                                                      & \multicolumn{3}{c|}{\textbf{CIFAR-100}}                                                     \\ \cline{2-10} 
\multicolumn{1}{|l|}{}                                            & \multicolumn{1}{c|}{\textit{Global}} & \multicolumn{1}{c|}{\textit{C-Gen}} & \textit{C-Spe} & \multicolumn{1}{c|}{\textit{Global}} & \multicolumn{1}{c|}{\textit{C-Gen}} & \textit{C-Spe} & \multicolumn{1}{c|}{\textit{Global}} & \multicolumn{1}{c|}{\textit{C-Gen}} & \textit{C-Spe} \\ \hline
\textbf{No Transfer}                                              & \multicolumn{1}{c|}{83.72}           & \multicolumn{1}{c|}{19.58}          & 98.80          & \multicolumn{1}{c|}{55.92}           & \multicolumn{1}{c|}{22.74}          & 78.11          & \multicolumn{1}{c|}{17.27}           & \multicolumn{1}{c|}{8.29}           & 29.23          \\ \hline
\textbf{FedAvg}                                                   & \multicolumn{1}{c|}{82.56}           & \multicolumn{1}{c|}{53.05}          & 98.14          & \multicolumn{1}{c|}{59.85}           & \multicolumn{1}{c|}{26.24}          & 79.66          & \multicolumn{1}{c|}{15.53}           & \multicolumn{1}{c|}{7.58}           & 26.67          \\ \hline
\textbf{Scaffold}                                                 & \multicolumn{1}{c|}{67.67}           & \multicolumn{1}{c|}{20.58}          & 98.60          & \multicolumn{1}{c|}{48.41}           & \multicolumn{1}{c|}{21.40}          & 74.25          & \multicolumn{1}{c|}{13.48}           & \multicolumn{1}{c|}{5.87}           & 20.52          \\ \hline
\textbf{MOON}                                                     & \multicolumn{1}{c|}{83.28}           & \multicolumn{1}{c|}{52.35}          & \textbf{99.30} & \multicolumn{1}{c|}{\textbf{60.83}}  & \multicolumn{1}{c|}{26.62}          & 79.55          & \multicolumn{1}{c|}{17.37}           & \multicolumn{1}{c|}{7.71}           & 27.62          \\ \hline
\textbf{FedDyn}                                                   & \multicolumn{1}{c|}{83.26}           & \multicolumn{1}{c|}{74.58}          & 95.58          & \multicolumn{1}{c|}{58.00}           & \multicolumn{1}{c|}{38.63}          & \textbf{79.85} & \multicolumn{1}{c|}{14.03}           & \multicolumn{1}{c|}{10.68}          & 29.78          \\ \hline
\textbf{\begin{tabular}[c]{@{}c@{}}CDKT\\ (Rep)\end{tabular}}     & \multicolumn{1}{c|}{84.88}           & \multicolumn{1}{c|}{\textbf{75.37}} & 96.74          & \multicolumn{1}{c|}{58.44}           & \multicolumn{1}{c|}{\textbf{40.59}} & 77.91          & \multicolumn{1}{c|}{17.75}           & \multicolumn{1}{c|}{\textbf{12.27}} & 31.34          \\ \hline
\textbf{\begin{tabular}[c]{@{}c@{}}CDKT\\ (Full)\end{tabular}}    & \multicolumn{1}{c|}{85.81}           & \multicolumn{1}{c|}{65.19}          & 98.14          & \multicolumn{1}{c|}{58.91}           & \multicolumn{1}{c|}{37.93}          & 73.74          & \multicolumn{1}{c|}{17.84}           & \multicolumn{1}{c|}{11.06}          & \textbf{32.80} \\ \hline
\textbf{\begin{tabular}[c]{@{}c@{}}CDKT\\ (RepFull)\end{tabular}} & \multicolumn{1}{c|}{\textbf{86.51}}  & \multicolumn{1}{c|}{69.70}          & 98.37          & \multicolumn{1}{c|}{59.09}           & \multicolumn{1}{c|}{39.88}          & 75.59          & \multicolumn{1}{c|}{\textbf{18.49}}  & \multicolumn{1}{c|}{11.69}          & 32.76          \\ \hline
\end{tabular}
}
}%
\end{center}
\end{table}

\begin{table}[]
\caption{\textcolor{black}{The median of F1 score (weighted) across three datasets over a range of $90$ to $100$ rounds in Fixed Users scenario.}}
\label{appendix_fixed_f1}
\textcolor{black}{
\begin{center}
\resizebox{14cm}{!}{%
\begin{tabular}{|c|ccc|ccc|ccc|}
\hline
\multicolumn{1}{|l|}{\multirow{2}{*}{}}                           & \multicolumn{3}{c|}{\textbf{Fashion-MNIST}}                                                 & \multicolumn{3}{c|}{\textbf{CIFAR-10}}                                                      & \multicolumn{3}{c|}{\textbf{CIFAR-100}}                                                     \\ \cline{2-10} 
\multicolumn{1}{|l|}{}                                            & \multicolumn{1}{c|}{\textit{Global}} & \multicolumn{1}{c|}{\textit{C-Gen}} & \textit{C-Spe} & \multicolumn{1}{c|}{\textit{Global}} & \multicolumn{1}{c|}{\textit{C-Gen}} & \textit{C-Spe} & \multicolumn{1}{c|}{\textit{Global}} & \multicolumn{1}{c|}{\textit{C-Gen}} & \textit{C-Spe} \\ \hline
\textbf{No Transfer}                                              & \multicolumn{1}{c|}{86.65}           & \multicolumn{1}{c|}{18.92}          & 98.72          & \multicolumn{1}{c|}{64.17}           & \multicolumn{1}{c|}{20.66}          & 75.82          & \multicolumn{1}{c|}{20.68}           & \multicolumn{1}{c|}{6.32}           & 22.56          \\ \hline
\textbf{FedAvg}                                                   & \multicolumn{1}{c|}{81.75}           & \multicolumn{1}{c|}{52.45}          & 98.62          & \multicolumn{1}{c|}{66.46}           & \multicolumn{1}{c|}{25.62}          & 78.07          & \multicolumn{1}{c|}{18.19}           & \multicolumn{1}{c|}{2.96}           & 22.62          \\ \hline
\textbf{Scaffold}                                                 & \multicolumn{1}{c|}{77.73}           & \multicolumn{1}{c|}{19.99}          & 97.85          & \multicolumn{1}{c|}{57.52}           & \multicolumn{1}{c|}{19.97}          & 72.31          & \multicolumn{1}{c|}{15.40}           & \multicolumn{1}{c|}{0.73}           & 19.66          \\ \hline
\textbf{MOON}                                                     & \multicolumn{1}{c|}{85.53}           & \multicolumn{1}{c|}{61.29}          & 98.63          & \multicolumn{1}{c|}{\textbf{69.03}}  & \multicolumn{1}{c|}{32.98}          & \textbf{80.60} & \multicolumn{1}{c|}{20.81}           & \multicolumn{1}{c|}{2.26}           & 23.69          \\ \hline
\textbf{FedDyn}                                                   & \multicolumn{1}{c|}{85.11}           & \multicolumn{1}{c|}{75.58}          & 96.36          & \multicolumn{1}{c|}{65.93}           & \multicolumn{1}{c|}{45.02}          & 79.66          & \multicolumn{1}{c|}{19.72}           & \multicolumn{1}{c|}{\textbf{12.82}} & 30.26          \\ \hline
\textbf{\begin{tabular}[c]{@{}c@{}}CDKT\\ (Rep)\end{tabular}}     & \multicolumn{1}{c|}{\textbf{87.04}}  & \multicolumn{1}{c|}{\textbf{81.61}} & 98.09          & \multicolumn{1}{c|}{67.15}           & \multicolumn{1}{c|}{\textbf{47.68}} & 77.12          & \multicolumn{1}{c|}{23.60}           & \multicolumn{1}{c|}{11.06}          & 32.28          \\ \hline
\textbf{\begin{tabular}[c]{@{}c@{}}CDKT\\ (Full)\end{tabular}}    & \multicolumn{1}{c|}{86.37}           & \multicolumn{1}{c|}{70.64}          & 98.40          & \multicolumn{1}{c|}{67.21}           & \multicolumn{1}{c|}{42.24}          & 74.40          & \multicolumn{1}{c|}{23.37}           & \multicolumn{1}{c|}{7.37}           & 31.93          \\ \hline
\textbf{\begin{tabular}[c]{@{}c@{}}CDKT\\ (RepFull)\end{tabular}} & \multicolumn{1}{c|}{86.69}           & \multicolumn{1}{c|}{66.14}          & \textbf{99.49} & \multicolumn{1}{c|}{68.15}           & \multicolumn{1}{c|}{43.59}          & 76.73          & \multicolumn{1}{c|}{\textbf{24.68}}  & \multicolumn{1}{c|}{9.71}           & \textbf{32.89} \\ \hline
\end{tabular}
}%
\end{center}
}
\end{table}

\begin{table}[t]
\caption{\textcolor{black}{The median of test accuracy across three datasets over a range of $90$ to $100$ rounds in Subset of Users scenario.}}
\label{appendix_subset}
\begin{center}
\resizebox{14cm}{!}{%
\textcolor{black}{
\begin{tabular}{|c|ccc|ccc|ccc|}
\hline
\multicolumn{1}{|l|}{\multirow{2}{*}{}}                           & \multicolumn{3}{c|}{\textbf{Fashion-MNIST}}                                                 & \multicolumn{3}{c|}{\textbf{CIFAR-10}}                                                      & \multicolumn{3}{c|}{\textbf{CIFAR-100}}                                                     \\ \cline{2-10} 
\multicolumn{1}{|l|}{}                                            & \multicolumn{1}{c|}{\textit{Global}} & \multicolumn{1}{c|}{\textit{C-Gen}} & \textit{C-Spe} & \multicolumn{1}{c|}{\textit{Global}} & \multicolumn{1}{c|}{\textit{C-Gen}} & \textit{C-Spe} & \multicolumn{1}{c|}{\textit{Global}} & \multicolumn{1}{c|}{\textit{C-Gen}} & \textit{C-Spe} \\ \hline
\textbf{No Transfer}                                              & \multicolumn{1}{c|}{78.19}           & \multicolumn{1}{c|}{19.09}          & \textbf{98.86} & \multicolumn{1}{c|}{52.17}           & \multicolumn{1}{c|}{17.14}          & \textbf{76.35} & \multicolumn{1}{c|}{16.47}           & \multicolumn{1}{c|}{3.04}           & 13.22          \\ \hline
\textbf{FedAvg}                                                   & \multicolumn{1}{c|}{69.44}           & \multicolumn{1}{c|}{65.33}          & 75.90          & \multicolumn{1}{c|}{48.83}           & \multicolumn{1}{c|}{41.69}          & 52.70          & \multicolumn{1}{c|}{12.25}           & \multicolumn{1}{c|}{10.81}          & 14.34          \\ \hline
\textbf{Scaffold}                                                 & \multicolumn{1}{c|}{68.00}           & \multicolumn{1}{c|}{62.36}          & 73.40          & \multicolumn{1}{c|}{40.76}           & \multicolumn{1}{c|}{37.97}          & 48.16          & \multicolumn{1}{c|}{12.29}           & \multicolumn{1}{c|}{10.89}          & 13.35          \\ \hline
\textbf{MOON}                                                     & \multicolumn{1}{c|}{70.05}           & \multicolumn{1}{c|}{66.53}          & 74.87          & \multicolumn{1}{c|}{53.05}           & \multicolumn{1}{c|}{46.82}          & 57.48          & \multicolumn{1}{c|}{16.79}           & \multicolumn{1}{c|}{\textbf{14.77}} & 18.61          \\ \hline
\textbf{FedDyn}                                                   & \multicolumn{1}{c|}{76.03}           & \multicolumn{1}{c|}{72.60}          & 79.62          & \multicolumn{1}{c|}{55.71}           & \multicolumn{1}{c|}{\textbf{47.84}} & 55.98          & \multicolumn{1}{c|}{16.37}           & \multicolumn{1}{c|}{14.59}          & 18.39          \\ \hline
\textbf{\begin{tabular}[c]{@{}c@{}}CDKT\\ (Rep)\end{tabular}}     & \multicolumn{1}{c|}{78.50}           & \multicolumn{1}{c|}{\textbf{69.70}} & 92.10          & \multicolumn{1}{c|}{55.67}           & \multicolumn{1}{c|}{38.33}          & 69.52          & \multicolumn{1}{c|}{16.72}           & \multicolumn{1}{c|}{7.41}           & 24.07          \\ \hline
\textbf{\begin{tabular}[c]{@{}c@{}}CDKT\\ (Full)\end{tabular}}    & \multicolumn{1}{c|}{79.22}           & \multicolumn{1}{c|}{62.86}          & 95.17          & \multicolumn{1}{c|}{\textbf{56.60}}  & \multicolumn{1}{c|}{38.02}          & 70.69          & \multicolumn{1}{c|}{17.33}           & \multicolumn{1}{c|}{7.76}           & 23.76          \\ \hline
\textbf{\begin{tabular}[c]{@{}c@{}}CDKT\\ (RepFull)\end{tabular}} & \multicolumn{1}{c|}{\textbf{79.94}}  & \multicolumn{1}{c|}{67.82}          & 94.96          & \multicolumn{1}{c|}{56.57}           & \multicolumn{1}{c|}{38.85}          & 69.22          & \multicolumn{1}{c|}{\textbf{17.60}}  & \multicolumn{1}{c|}{10.32}          & \textbf{25.15} \\ \hline
\end{tabular}
}
}%
\end{center}
\end{table}

\begin{table}[t]
\caption{\textcolor{black}{The median of F1 score (weighted) across three datasets over a range of $90$ to $100$ rounds in Subset of Users scenario.}}
\label{appendix_subset_F1}
\begin{center}
\resizebox{14cm}{!}{%
\textcolor{black}{
\begin{tabular}{|c|ccc|ccc|ccc|}
\hline
\multicolumn{1}{|l|}{\multirow{2}{*}{}}                           & \multicolumn{3}{c|}{\textbf{Fashion-MNIST}}                                                 & \multicolumn{3}{c|}{\textbf{CIFAR-10}}                                                      & \multicolumn{3}{c|}{\textbf{CIFAR-100}}                                                     \\ \cline{2-10} 
\multicolumn{1}{|l|}{}                                            & \multicolumn{1}{c|}{\textit{Global}} & \multicolumn{1}{c|}{\textit{C-Gen}} & \textit{C-Spe} & \multicolumn{1}{c|}{\textit{Global}} & \multicolumn{1}{c|}{\textit{C-Gen}} & \textit{C-Spe} & \multicolumn{1}{c|}{\textit{Global}} & \multicolumn{1}{c|}{\textit{C-Gen}} & \textit{C-Spe} \\ \hline
\textbf{No Transfer}                                              & \multicolumn{1}{c|}{85.65}           & \multicolumn{1}{c|}{18.35}          & \textbf{97.48} & \multicolumn{1}{c|}{64.84}           & \multicolumn{1}{c|}{11.04}          & 60.90          & \multicolumn{1}{c|}{21.57}           & \multicolumn{1}{c|}{1.75}           & 8.21           \\ \hline
\textbf{FedAvg}                                                   & \multicolumn{1}{c|}{73.91}           & \multicolumn{1}{c|}{70.76}          & 80.54          & \multicolumn{1}{c|}{61.54}           & \multicolumn{1}{c|}{53.68}          & 63.28          & \multicolumn{1}{c|}{12.50}           & \multicolumn{1}{c|}{10.90}          & 13.76          \\ \hline
\textbf{Scaffold}                                                 & \multicolumn{1}{c|}{75.51}           & \multicolumn{1}{c|}{78.79}          & 87.59          & \multicolumn{1}{c|}{51.34}           & \multicolumn{1}{c|}{46.96}          & 53.77          & \multicolumn{1}{c|}{14.04}           & \multicolumn{1}{c|}{\textbf{12.96}} & 14.22          \\ \hline
\textbf{MOON}                                                     & \multicolumn{1}{c|}{74.53}           & \multicolumn{1}{c|}{71.70}          & 82.09          & \multicolumn{1}{c|}{65.00}           & \multicolumn{1}{c|}{\textbf{58.56}} & 66.88          & \multicolumn{1}{c|}{18.85}           & \multicolumn{1}{c|}{12.62}          & 15.06          \\ \hline
\textbf{FedDyn}                                                   & \multicolumn{1}{c|}{81.33}           & \multicolumn{1}{c|}{\textbf{79.19}} & 83.91          & \multicolumn{1}{c|}{64.32}           & \multicolumn{1}{c|}{56.09}          & 65.18          & \multicolumn{1}{c|}{18.54}           & \multicolumn{1}{c|}{12.59}          & 15.87          \\ \hline
\textbf{\begin{tabular}[c]{@{}c@{}}CDKT\\ (Rep)\end{tabular}}     & \multicolumn{1}{c|}{86.41}           & \multicolumn{1}{c|}{76.51}          & 95.41          & \multicolumn{1}{c|}{65.69}           & \multicolumn{1}{c|}{48.86}          & \textbf{75.96} & \multicolumn{1}{c|}{22.61}           & \multicolumn{1}{c|}{7.17}           & \textbf{21.73} \\ \hline
\textbf{\begin{tabular}[c]{@{}c@{}}CDKT\\ (Full)\end{tabular}}    & \multicolumn{1}{c|}{86.39}           & \multicolumn{1}{c|}{70.30}          & 96.40          & \multicolumn{1}{c|}{\textbf{67.07}}  & \multicolumn{1}{c|}{46.68}          & 74.67          & \multicolumn{1}{c|}{\textbf{22.86}}  & \multicolumn{1}{c|}{7.12}           & 18.76          \\ \hline
\textbf{\begin{tabular}[c]{@{}c@{}}CDKT\\ (RepFull)\end{tabular}} & \multicolumn{1}{c|}{\textbf{86.46}}  & \multicolumn{1}{c|}{75.51}          & 96.29          & \multicolumn{1}{c|}{66.63}           & \multicolumn{1}{c|}{51.75}          & 74.27          & \multicolumn{1}{c|}{22.67}           & \multicolumn{1}{c|}{7.52}           & 18.69          \\ \hline
\end{tabular}
}
}%
\end{center}
\end{table}

\begin{table}[t]
\textcolor{black}{
\begin{center}
\caption{\textcolor{black}{The median of test accuracy of \CDKT on Fashion-MNIST dataset in different settings of clients.}}
\label{scalability_fmnist}
\resizebox{14cm}{!}{%
\begin{tabular}{|l|cc|cc|cc|}
\hline
\multirow{3}{*}{}       & \multicolumn{2}{c|}{\textbf{50 clients, sample fraction = 0.2}} & \multicolumn{2}{c|}{\textbf{50 clients, sample fraction = 0.5}} & \multicolumn{2}{c|}{\textbf{100 clients, sample fraction = 0.2}} \\ \cline{2-7} 
                        & \multicolumn{2}{c|}{100 rounds}                                 & \multicolumn{2}{c|}{100 rounds}                                 & \multicolumn{2}{c|}{200 rounds}                                  \\ \cline{2-7} 
                        & \multicolumn{1}{p{3cm}|}{\centering\textit{Global}}      & \textit{C-Per}      & \multicolumn{1}{p{3cm}|}{\centering\textit{Global}}       & \textit{C-Per}      & \multicolumn{1}{p{3cm}|}{\centering\textit{Global}}        & \textit{C-Per}      \\ \hline
\textbf{CDKT (Rep)}     & \multicolumn{1}{c|}{78.50}                & 80.90               & \multicolumn{1}{c|}{77.83}                & 82.42               & \multicolumn{1}{c|}{79.23}                 & 79.02               \\ \hline
\textbf{CDKT (Full)}    & \multicolumn{1}{c|}{79.22}                & 79.02               & \multicolumn{1}{c|}{78.91}                & 82.04               & \multicolumn{1}{c|}{79.33}                 & 79.41               \\ \hline
\textbf{CDKT (RepFull)} & \multicolumn{1}{c|}{\textbf{79.94}}       & \textbf{81.39}      & \multicolumn{1}{c|}{\textbf{78.92}}       & \textbf{83.32}      & \multicolumn{1}{c|}{\textbf{80.23}}        & \textbf{79.69}      \\ \hline
\end{tabular}
}%
\end{center}
}
\end{table}

\begin{table}[t]
\textcolor{black}{
\caption{\textcolor{black}{The median of test accuracy of \CDKT on CIFAR-10 dataset in different settings of clients.}}
\label{scalability_cifar10}
\begin{center}
\resizebox{14cm}{!}{%
\begin{tabular}{|l|cc|cc|cc|}
\hline
\multirow{3}{*}{}       & \multicolumn{2}{c|}{\textbf{50 clients, sample fraction = 0.2}} & \multicolumn{2}{c|}{\textbf{50 clients, sample fraction = 0.5}} & \multicolumn{2}{c|}{\textbf{100 clients, sample fraction = 0.2}} \\ \cline{2-7} 
                        & \multicolumn{2}{c|}{100 rounds}                                 & \multicolumn{2}{c|}{100 rounds}                                 & \multicolumn{2}{c|}{200 rounds}                                  \\ \cline{2-7} 
                        & \multicolumn{1}{p{3cm}|}{\centering\textit{Global}}      & \textit{C-Per}      & \multicolumn{1}{p{3cm}|}{\centering\textit{Global}}      & \textit{C-Per}      & \multicolumn{1}{p{3cm}|}{\centering\textit{Global}}       & \textit{C-Per}      \\ \hline
\textbf{CDKT (Rep)}     & \multicolumn{1}{c|}{55.67}                & 53.92               & \multicolumn{1}{c|}{53.72}                & 57.66               & \multicolumn{1}{c|}{54.06}                 & 47.27               \\ \hline
\textbf{CDKT (Full)}    & \multicolumn{1}{c|}{\textbf{56.60}}       & \textbf{54.35}      & \multicolumn{1}{c|}{54.78}                & 56.05               & \multicolumn{1}{c|}{\textbf{56.12}}        & 47.47               \\ \hline
\textbf{CDKT (RepFull)} & \multicolumn{1}{c|}{56.57}                & 54.03               & \multicolumn{1}{c|}{\textbf{54.93}}       & \textbf{57.86}      & \multicolumn{1}{c|}{55.72}                 & \textbf{49.42}      \\ \hline
\end{tabular}
}%
\end{center}
}
\end{table}

\begin{table}[t]
\textcolor{black}{
\caption{\textcolor{black}{The median of test accuracy of \CDKT on CIFAR-100 dataset in different settings of clients.}}
\label{scalability_cifar100}
\begin{center}
\resizebox{14cm}{!}{%
\begin{tabular}{|l|cc|cc|cc|}
\hline
\multirow{3}{*}{}       & \multicolumn{2}{c|}{\textbf{50 clients, sample fraction = 0.2}} & \multicolumn{2}{c|}{\textbf{50 clients, sample fraction = 0.5}} & \multicolumn{2}{c|}{\textbf{100 clients, sample fraction = 0.2}} \\ \cline{2-7} 
                        & \multicolumn{2}{c|}{100 rounds}                                 & \multicolumn{2}{c|}{100 rounds}                                 & \multicolumn{2}{c|}{200 rounds}                                  \\ \cline{2-7} 
                        & \multicolumn{1}{p{3cm}|}{\centering\textit{Global}}      & \textit{C-Per}      & \multicolumn{1}{p{3cm}|}{\centering\textit{Global}}      & \textit{C-Per}      & \multicolumn{1}{p{3cm}|}{\centering\textit{Global}}       & \textit{C-Per}      \\ \hline
\textbf{CDKT (Rep)}     & \multicolumn{1}{c|}{16.72}                & 15.74               & \multicolumn{1}{c|}{16.92}                & 18.58               & \multicolumn{1}{c|}{17.17}                 & 16.26               \\ \hline
\textbf{CDKT (Full)}    & \multicolumn{1}{c|}{17.33}                & 15.76               & \multicolumn{1}{c|}{17.58}                & \textbf{20.80}      & \multicolumn{1}{c|}{17.95}                 & \textbf{16.47}      \\ \hline
\textbf{CDKT (RepFull)} & \multicolumn{1}{c|}{\textbf{17.60}}       & \textbf{17.74}      & \multicolumn{1}{c|}{\textbf{17.60}}       & 18.78               & \multicolumn{1}{c|}{\textbf{17.97}}        & 16.29               \\ \hline
\end{tabular}
}%
\end{center}
}
\end{table}


\end{document}